\documentclass[10pt,twocolumn,letterpaper]{article}

\usepackage{cvpr}      

\usepackage{times}
\usepackage{epsfig}
\usepackage{graphicx}
\usepackage{amsmath}
\usepackage{amssymb}
\usepackage{booktabs}
\usepackage{comment}
\usepackage{enumitem}
\usepackage[misc]{ifsym}
\usepackage[american]{babel}

\newcommand{\argmin}{\mathop{\arg\min}}

\newcommand{\x}{{\bf x}}
\newcommand{\aaa}{{\bf a}}

\newcommand{\p}{{\bf p}}

\newcommand{\z}{{\bf z}}

\newcommand{\y}{{\bf y}}

\newcommand{\name}{{\sc DiscoverNet}}

\usepackage[pagebackref,breaklinks,colorlinks]{hyperref}

\usepackage[capitalize]{cleveref}
\crefname{section}{Sec.}{Secs.}
\Crefname{section}{Section}{Sections}
\Crefname{table}{Table}{Tables}
\crefname{table}{Tab.}{Tabs.}

\def\cvprPaperID{5564} 
\def\confName{CVPR}
\def\confYear{2022}

\begin{document}

\title{Identifying Ambiguous Similarity Conditions via Semantic Matching}

\author{Han-Jia Ye\qquad Yi Shi\qquad De-Chuan Zhan\\
State Key Laboratory for Novel Software Technology, Nanjing University
\\
{\tt\small \{yehj, shiy, zhandc\}@lamda.nju.edu.cn}
}

\maketitle

\begin{abstract}
Rich semantics inside an image result in its ambiguous relationship with others, \ie, two images could be similar in one condition but dissimilar in another. 
Given triplets like ``aircraft'' is similar to ``bird'' than ``train'', Weakly Supervised Conditional Similarity Learning (WS-CSL) learns multiple embeddings to match semantic conditions {\em without explicit condition labels} such as ``can fly''. 
However, similarity relationships in a triplet are uncertain except {\em providing a condition}. For example, the previous comparison becomes invalid once the conditional label changes to ``is vehicle''.
To this end, we introduce a novel evaluation criterion by predicting the comparison's correctness after assigning the learned embeddings to their optimal conditions, which measures how much WS-CSL could cover latent semantics as the supervised model.
Furthermore, we propose the Distance Induced Semantic COndition VERification Network ({\name}), which characterizes the instance-instance and triplets-condition relations in a ``decompose-and-fuse'' manner. 
To make the learned embeddings cover all semantics, {\name} utilizes a set module or an additional regularizer over the correspondence between a triplet and a condition.
{\name} achieves state-of-the-art performance on benchmarks like UT-Zappos-50k and Celeb-A w.r.t. different criteria.
\end{abstract}

\section{Introduction}\label{sec:intro}
Learning embeddings (a.k.a. representations) from data benefits machine learning and visual recognition systems~\cite{davis-et-al-icml-2007,weinberger09distance,amid2015MVTE,Schroff2015FaceNet,Ye2016What,feng2020uncertainty,Opitz2020Deep,Ye2020Learning,zhou2022forward}. 
Side information such as triplets~\cite{Song2016Deep,Sohn2016Improved,Manmatha2017Sampling,Elezi2020Group,Duan2020Deep} indicates the comparison relationship of objects, from which the embedding pulls visually similar objects close while pushing dissimilar ones away.

\begin{figure}[t]
\vspace{-3mm}
	\centering
	\begin{minipage}[c]{\linewidth}
	\includegraphics[width=\linewidth]{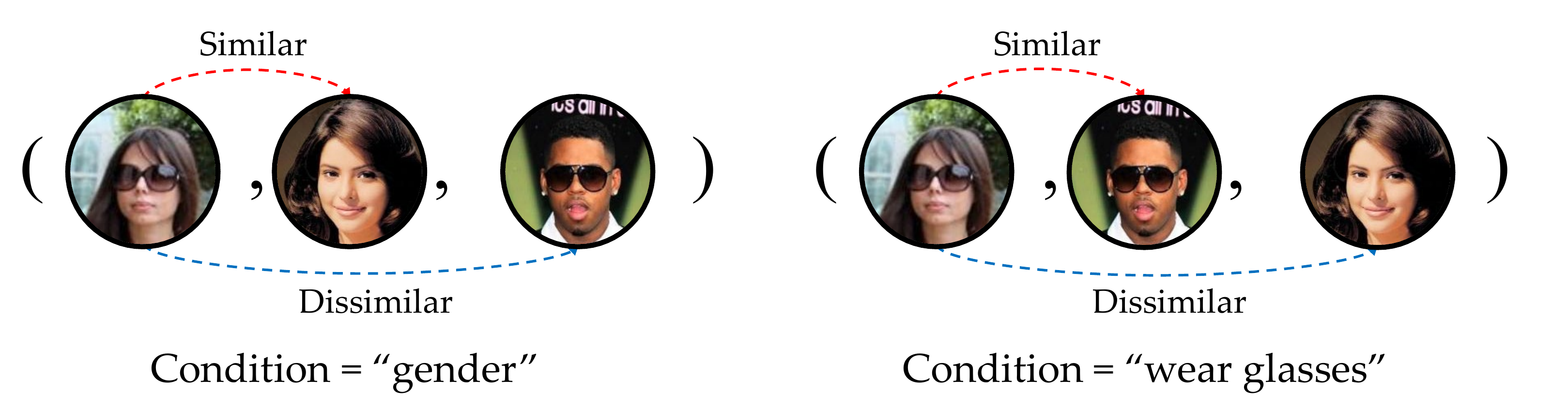}\\
	\vspace{-5mm}
    \end{minipage}
    \begin{minipage}[c]{0.48\linewidth}
	\includegraphics[width=\linewidth]{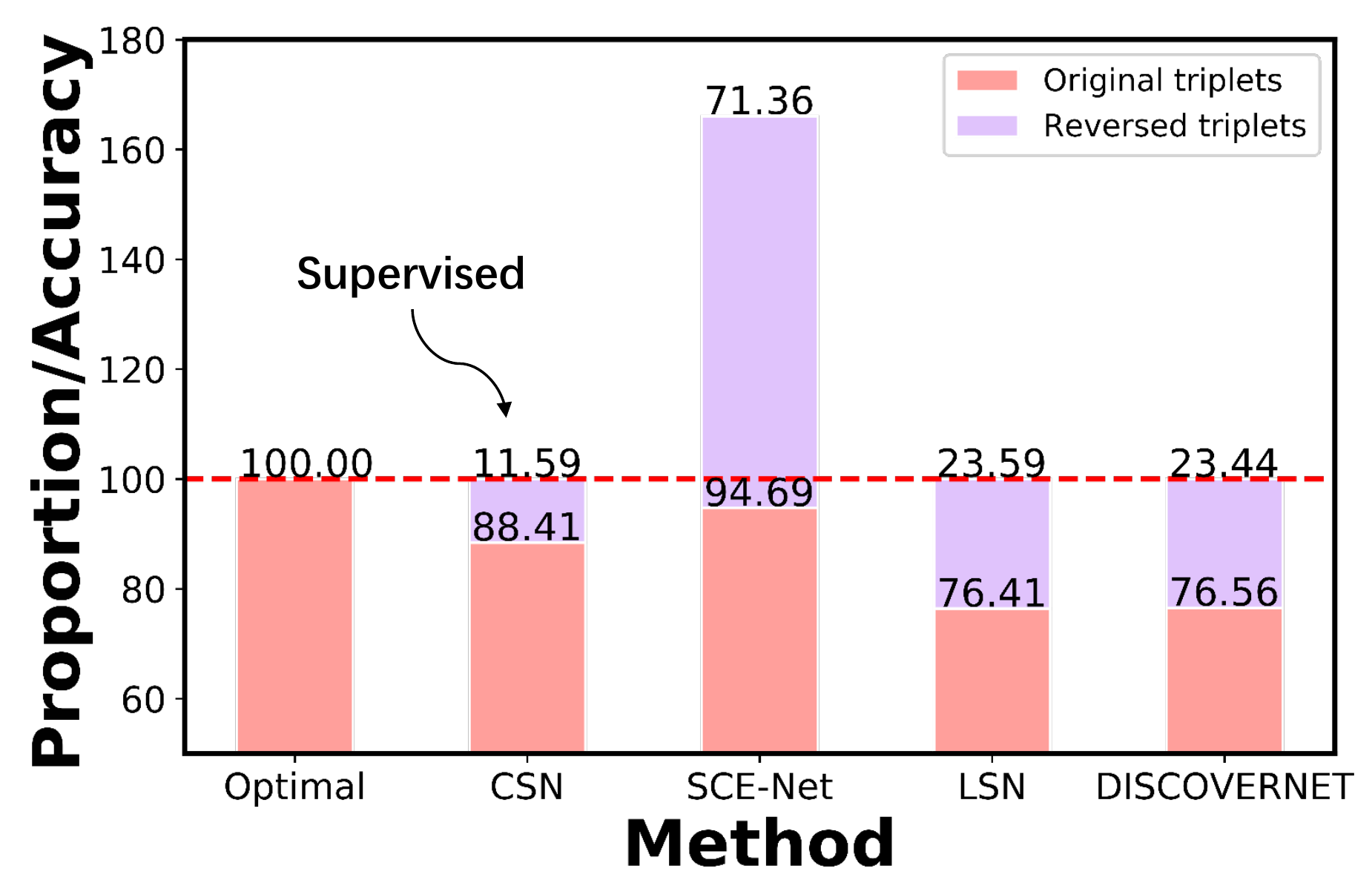}\\
    \end{minipage} 
    \begin{minipage}[c]{0.48\linewidth}
	\includegraphics[width=\linewidth]{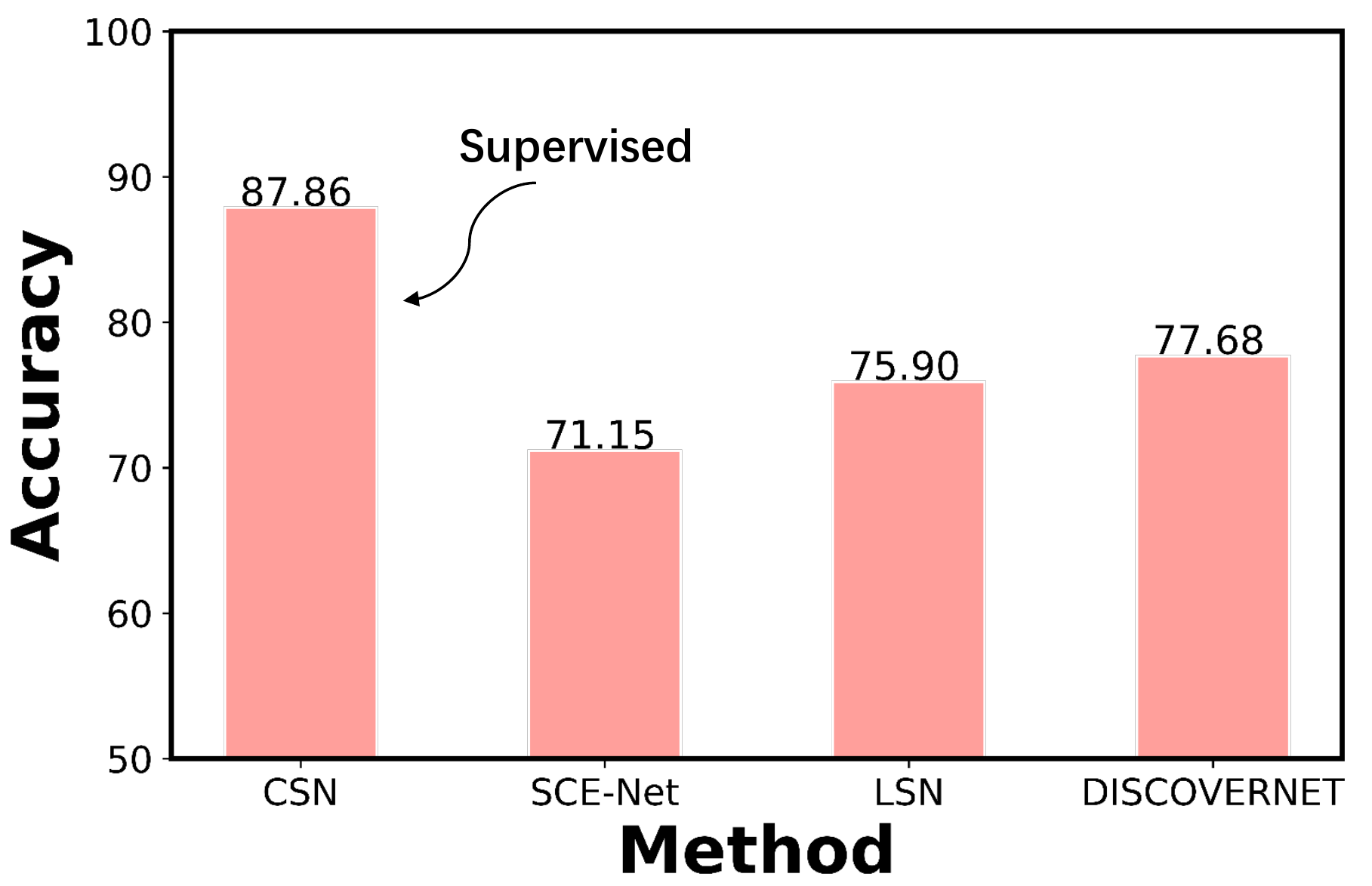}\\
    \end{minipage}
	\vspace{-6mm}
	\caption{Upper: Two triplets with the same set of instances could be {\em both} meaningful when we measure similarity from different conditions.
	Lower Left: Given original (correct) triplets and their reversed variants (invalid w.r.t. the same similarity condition) on UT-Zappos-50k, we compute the proportion a model predicts them as valid ones. Last three are WS-CSL methods.
	Lower right: Our proposed criterion avoids the issue of reversed triplets naturally, and our {\name} outperforms other WS-CSL methods.}
	\label{fig:teaser}
	\vspace{-5mm}
\end{figure}

The linkage between objects conveys rich information about the object itself as well as its relationship with others, which becomes ambiguous when the similarity is measured from different perspectives. 
As illustrated in Fig.~\ref{fig:teaser} (upper), (a) female with glasses, (b) female without glasses, and (c) male with glasses are organized in triplets.\footnote{In a valid triplet $(a,b,c)$, $(a,b)$ is more similar than $(a,c)$.} We think (a) and (b) are more similar when we measure based on ``gender''. In contrast, we also treat (a) and (c) as neighbors since they ``wear glasses''. 
Since one embedding space outputs fixed relationships between instances, learning multiple embeddings facilitates discovering rich semantics.

Given triplets associated with their condition labels, indicating under what kind of similarity the comparisons are made, Conditional Similarity Learning~(CSL) learns multiple embeddings to cover latent semantic~\cite{Veit2017Conditional,Wang2017Contextual}. 
During the evaluation, CSL predicts whether a triplet is meaningful or not under a specified condition.
Although supervised CSL has been successfully applied in various applications~\cite{Lee2018Stacked,Plummer2018Conditional,Vasileva2018Learning,Lin2020Fashion}, labeling conditions introduces additional costs. 
As in a recommendation system, users may click relevant items (label item-wise similarities) based on particular preferences, and we only collect diverse comparison relationships {\em without explicit condition labels}. \cite{Tan2019Learning,Nigam2019Towards} propose {\em Weakly Supervised}-CSL (WS-CSL), where multiple embeddings are learned with triplets and the model is {\em unaware of} their corresponding conditions. 

Current WS-CSL borrows the evaluation protocol from supervised CSL~\cite{Veit2017Conditional}, which checks the validness of a triplet but {\em neglects the specified condition}. For example, we predict whether ``aircraft'' is similar to ``bird'' than ``train'' conditioned on ``can fly'' in the supervised scenario, but ask the model to predict the correctness of the triplet {\em free from the condition} in WS-CSL.
Since a triplet could be ambiguous, WS-CSL may focus on {\em weighting those embeddings to explain the triplet instead of learning semantically conditional embeddings corresponding to the ground-truth}.

We demonstrate the challenge with the following experiment. 
After removing all condition labels of correct triplets, we compute the proportion a WS-CSL model predicts those triplets as valid ones via its learned embeddings, which equals the ``accuracy'' used in WS-CSL evaluation. 
Then we {\em reverse} those triplets --- changing the order of the second and the third items, which makes them invalid under previous conditions. We further compute the valid ratio over reversed triplets. Results in Fig.~\ref{fig:teaser} (lower left) show an optimal supervised model has 100\% and 0\% proportion/accuracy in two cases. A WS-CSL method SCE-Net~\cite{Tan2019Learning} has a higher proportion than the supervised CSN~\cite{Veit2017Conditional} on original triplets as well as a high proportion on reversed ones, which means it treats both original and reversed triplets as valid. 
The diverse results between WS-CSL and its supervised counterparts indicate the ``accuracy'' is biased towards predicting all triplets as valid, and the correctness of a triplet without a specified condition is meaningless.
To check the coverage of all semantics, we propose to measure the comparison ability of multiple learned embeddings after {\em assigning them to target conditions}.

We also propose Distance Induced Semantic COndition VERification Network~(\name) to balance the ability of {\em comparison prediction} and {\em semantic coverage}. 
{\name} works in a ``decompose-and-fuse'' manner, which identifies similarity conditions and captures the ambiguous relationship with discriminative embeddings. 

We aim to ensure the learned multiple embeddings in WS-CSL are able to reveal the similarities in the target conditions as much as the supervised methods.
In {\name}, we achieve the goal from two aspects. First, we use a set module to map various triplets with the same set of instances to one condition, avoiding messy training update signals.
On the other hand, we add a regularizer to force selecting different conditional embeddings when a model predicts both a triplet and its reversed one as valid.
{\name} demonstrates higher performance on benchmarks like UT-Zappos-50k in our newly proposed criterion, which is shown in Fig.~\ref{fig:teaser} (lower right). The code is available at \url{https://github.com/shiy19/DiscoverNet}.

Our contributions could be summarized as:
\begin{itemize}
	\item We point out the challenges in WS-CSL evaluation and design a novel criterion.
	\item Based on the proposed {\name}, we improve the quality of learned embeddings and match the ground-truth semantics from two aspects.
	\item {\name} can identify latent rich conditions and works better than others in our criterion. 
\end{itemize}

\section{Related Work}\label{sec:related}
\noindent\textbf{Metric Learning.} Learning embeddings from data attracts lots of attention in machine learning and computer vision. The learned embeddings encode the relationship between objects well and facilitate downstream tasks~\cite{chechik2010large,Schroff2015FaceNet,Song2016Deep,Hsieh2017Collaborative}.
With the guidance of the comparison relationship between object pairs~\cite{davis-et-al-icml-2007}, triplets~\cite{weinberger09distance,Ye2020Learning}, and higher-order statistics~\cite{law2016learning}, visually similar instances are pulled together, and visually dissimilar ones are pushed away.
Types of loss functions are proposed~\cite{Bell2015Learning,Song2016Deep,Song2017Deep,Opitz2020Deep,Manmatha2017Sampling,Elezi2020Group,Duan2020Deep} to take full advantage of the similarity comparisons between instances in a mini-batch, \eg, the triplet loss~\cite{Schroff2015FaceNet} and N-Pair loss~\cite{Sohn2016Improved}. 

\noindent\textbf{Conditional Similarity Learning (CSL).} Different from measuring all linkages with a single metric, the relationship between objects could be measured from diverse aspects (under different conditions)~\cite{Memisevic2004Multiple,Maaten2012Visualizing,NIPS2013_SCA}.
In~\cite{Veit2017Conditional,Yu2017Semantic}, CSL is investigated by associating conditions with image attributes, and multiple diverse embeddings could be derived from feature masks to capture the semantic of various conditions.
CSL has been applied in applications like image-text classification~\cite{Liu2019Neural,Nigam2019Towards}, 
fashion retrieval~\cite{Tan2019Learning, Hou2021Learning, Dong2021Fine}, zero-shot learning~\cite{Zhang2015Zero} and video grounding~\cite{Shi2019Not}. 
Condition labels during training link a particular embedding with a condition, and the learned embeddings are evaluated by predicting the validness of a triplet {\em under a certain condition}.

\noindent\textbf{Weakly supervised CSL.}
Explicit condition labels are unavailable in some cases, \eg, we get a comparison tuple once a user selects an item than others without any knowledge of his/her preference (condition). 
The weakly supervised CSL, \ie, learning conditional embeddings without condition annotations is investigated in~\cite{amid2015MVTE}, where a model infers conditions given a triplet and then decides the right embedding to use for training and deployment. 
\cite{Tan2019Learning} emphasizes the comparison ability of the fused embedding, while \cite{Ye2016What,Nigam2019Towards} select one metric from multiple candidates to explain a triplet. 
Using the same evaluation protocols as CSL, weakly supervised CSL can get higher performance than supervised CSL without involving the ground-truth condition labels. We analyze the protocol and point out its drawbacks of semantic coverage. We propose a new criterion to reveal the difference between the quality of the learned embedding and its supervised counterpart. 
We also propose {\name} to trade-off the diversified embedding and semantic fusion in a ``decompose and fuse'' manner. A set module and a semantic regularizer are discussed to make {\name} cover all target conditions.

\section{Notations and Background}\label{sec:background}
We organize comparisons into triplets, \ie, $\mathcal{T} = \{\tau=(\x, \y, \z)\}$. 
For three items in $\tau$: the anchor $\x\in\mathbb{R}^D$ has a similar target neighbor $\y\in\mathbb{R}^D$, and an impostor $\z\in\mathbb{R}^D$ is dissimilar with $\x$. 
Similarity in $\tau$ could be determined by categories of instances.
The embedding, a.k.a., feature extractor, $\phi:\mathbb{R}^D\rightarrow\mathbb{R}^d$ projects an instance $\x$ into a latent space, whose distance with $\y$ is
{\begin{equation}
\mathbf{Dis}_L^2(\phi(\x), \;\phi(\y)) = \|L^\top(\phi(\x) - \phi(\y))\|_2^2\;.\label{eq:1}
\end{equation}}$L\in\mathbb{R}^{d\times d'}$ is a projection.
We implement $\phi$ with deep neural network, and learn embedding by minimizing the violation of comparisons over $|\mathcal{T}|$ triplets. Define the validness of a triplet $\tau$ by comparing the distances between ``anchor-impostor'' and ``anchor-target neighbor'', \ie,
{\begin{equation}
\mathbf{Diff}_\tau = \mathbf{Dis}_L^2(\phi(\x), \;\phi(\z)) - \mathbf{Dis}_L^2(\phi(\x), \;\phi(\y))\;.
\end{equation}}If the anchor has a larger distance with the impostor than with the target neighbor, the embedding is consistent with the relationship in $\tau$. Given a loss $\ell(\cdot)$, \eg, the margin loss or the logistic loss, we optimize $\min_{\phi, L}\;\sum_{\tau\in\mathcal{T}}\; \ell\left(\mathbf{Diff}_\tau - \gamma\right)$. Then $\mathbf{Diff}_\tau$ would be larger than the margin $\gamma$, so that anchor and the target neighbor are pulled while the impostor would be pushed away. 

\paragraph{Conditional similarity learning~(CSL).} Comparisons in triplet $\tau$ may vary across environments. For example, when we ask which image in a candidate set is close to a given anchor, different workers may measure the similarities from their own aspects.
In CSL, we associate a condition label $k\in\{1,\ldots,K\}$ with each triplet, \ie, $\mathcal{T} = \{\tau=(\x,\y,\z,k)\}$, indicating based on the $k$-th condition the comparison in $\tau$ is made. 
One embedding $\phi$ fails to capture the diverse relationship
when comparisons with the same set of instances are opposite, \eg, $\tau_1=(\x,\y,\z,k)$ vs. $\tau_2=(\x,\z,\y,k')$. 
CSL extends $\phi$, $L$, and $\mathbf{Diff}_\tau$ to $\phi_k$, $L_k$, and $\mathbf{Diff}^k_\tau$, respectively. $K$ embeddings $\Psi_K$
covering $K$ conditions are optimized:
{\begin{equation}
\min_{\Psi_K=\{\psi_k=L_k\circ\phi_k\}_{k=1}^K}\;\sum_{\tau\in\mathcal{T}} \ell\left(\sum_{k'=1}^K I[k'=k] \left[\mathbf{Diff}^{k'}_\tau\right] - \gamma\right)\;.\label{eq:csl_obj}
\end{equation}}$I[\cdot]$ outputs 1 when the input is true and 0 otherwise. In Eq.~\ref{eq:csl_obj}, only the distance corresponding to condition $k$ of the triplet $\tau$ is activated. 
In evaluation, $\psi_k$ is used to check whether a triplet from the $k$-th condition is valid or not.

\paragraph{Weakly supervised CSL~(WS-CSL).}
Due to additional annotation costs and ambiguity of condition labels, the triplet-wise condition labels $\{k\}$ are {\em unknown} in WS-CSL.
The model needs to {\em infer} condition labels and activates the corresponding embeddings to capture triplet's characteristic. The target of WS-CSL is to learn a set of embeddings $\Psi_K$, which is able to capture those semantics of target conditions as much as the supervised CSL.

\section{Evaluation of Weakly Supervised CSL}\label{sec:evaluation}
Current WS-CSL follows CSL to measure $\psi_k$ by predicting whether three instances could form a triplet {\em without a condition label}. We analyze the issues of the current criterion and propose our new evaluation protocol.

\begin{figure}[!t]
	\centering
	\begin{minipage}[h]{\linewidth}
		\centering
		\includegraphics[width=\linewidth]{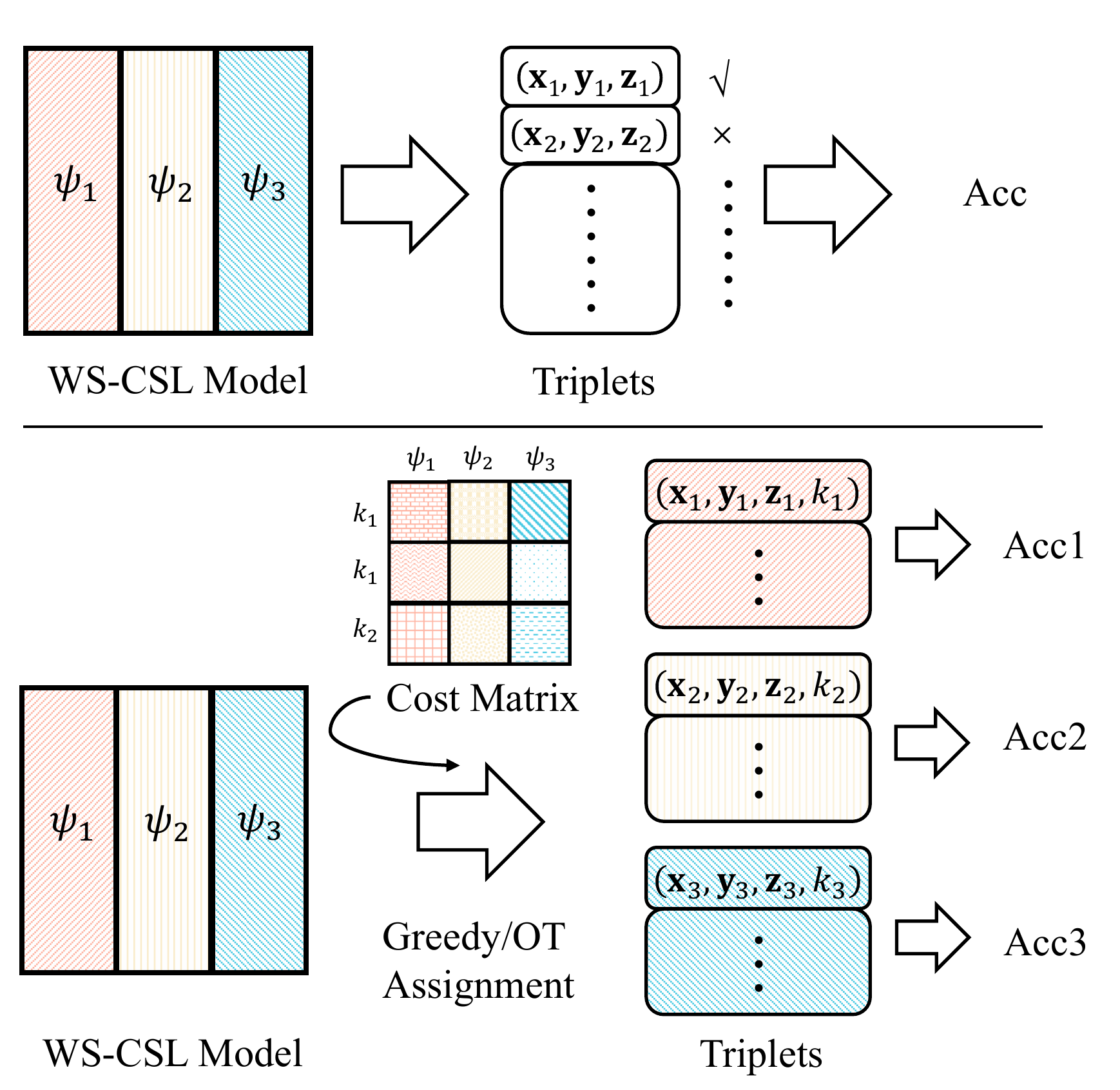}
	\end{minipage}
	\vspace{-3mm}
	\caption{Comparison between previous (upper) and our proposed (lower) WS-CSL criterion. Instead of predicting the validness of triplets indistinguishably, We align the learned embeddings with target conditions and compute condition-specific accuracy.}\label{fig:evaluation}
	\vspace{-5mm}
\end{figure}

\subsection{Analysis of Supervised CSL Evaluation}
In supervised CSL evaluation, a model is asked to determine whether a triplet $\tau_1=(\x,\y,\z,k)$ is valid or not given the condition $k$. 
In detail, we get $\mathbf{Diff}^{k}_\tau$ with the corresponding conditional embedding $\psi_k$ and predict $\tau_1$ as valid if $\mathbf{Diff}^{k}_\tau > 0$.
Since $\tau_1=(\x,\y,\z,k)$ and $\tau_2=(\x,\z,\y,k)$ could not co-exist, we sample the same number of valid triplets from each condition in evaluation. 
The average accuracy (proportion of triplets predicted as correct) over them reveals the quality of a CSL model.

Since the ability of comparison prediction is important, previous WS-CSL methods follow this protocol {\em without using the test-time condition labels}. In other words, we predict whether a triplet $\tau_3=(\x,\y,\z)$ is correct without its condition label. 
So in this case, a WS-CSL model predicts the validness of a triplet with {\em all learned embeddings}.

There are two issues when evaluating with the supervised protocol directly.
First, a model tends to find a condition to explain the triplet, which predicts all triplets as true in most cases, \eg, even for reversed triplets (demonstrated in Fig.~\ref{fig:teaser}). Specifically, given a valid $\tau_1=(\x,\y,\z,k)$, we construct $\tau_2=(\x,\z,\y,k)$, and ask a WS-CSL model whether $\tau_4=(\x,\z,\y)$ is valid or not. Since $\tau_2$ is invalid under the same condition $k$, an optimal supervised model will predict $\tau_1$ as true and $\tau_2$ as false. However, a WS-CSL method SCE-Net~\cite{Tan2019Learning} predicts most original ($\tau_3$) and reversed ($\tau_4$) triplets as valid, which makes the evaluation biased.

Moreover, the current evaluation predicts the validness of the triplet while neglecting from which condition the WS-CSL model determines the relationship. We'd like a WS-CSL model works similarly to a supervised one, so not only the fusion of conditional embeddings $\Psi_K$ should cover all target semantics, but also the behavior of each $\psi_k$ reveals the relationship w.r.t. a specific condition.
Therefore, we propose a new criterion to meet the previous requirements.

\subsection{Condition Alignment for WS-CSL Evaluation}\label{subsec:evaluation}
We claim that predicting the validness of a triplet is only meaningful {\em given a specific condition}.
Based on the ground-truth conditions and the validness of triplets {\em during the evaluation}, we propose a {\em two-step} new criterion. First, we map target conditions to WS-CSL embeddings $\Psi_K$. Then, we evaluate triplets accuracy with corresponding $\psi_k$ as in the supervised scenario, which reveals how much a WS-CSL model covers the target conditions as the supervised model.

We assume there is the same number of embeddings and conditions.\footnote{Our analysis could be extended when their numbers do not match.} Given triplets from the $k$-th condition for evaluation, we compute the triplet prediction accuracy with $\{\psi_{k'}\}_{k'=1}^K$. We collect accuracy for all conditions and form a cost matrix $C\in\mathbb{R}^{K\times K}$, whose element $C_{k'k}$ is the error (100 minus the accuracy) using $k'$-th embedding to predict the triplets from the $k$-th condition. The alignment could be obtained from $C$ with the following two strategies.

\noindent{\bf Greedy Alignment.} We use a greedy strategy to find the most suitable embedding, \ie, $\argmin_{k'}C_{k'k}$, for a condition $k$. So one embedding may handle multiple conditions.

\noindent{\bf OT Alignment.} We optimize the following Optimal Transport~(OT) objective~\cite{villani2008optimal} to obtain a mapping $T\in\mathbb{R}^{K\times K}$ from the embedding set to the condition set
{
\begin{equation}
\min_{T\ge0} \langle T, C \rangle\qquad {\rm s.t.}\quad T{\mathbf 1} = \frac{1}{K}{\mathbf 1}, \;T^\top{\mathbf 1} = \frac{1}{K}{\mathbf 1}\;.\label{eq:ot}
\end{equation}}Eq.~\ref{eq:ot} minimizes the total cost when we use an embedding to predict triplets from another condition. We set the marginal distribution of the transportation $T$ as uniform. By minimizing Eq.~\ref{eq:ot} we obtain $T$ as a map. Element $T_{k'k}$ reveals how much the $k'$-th embedding is related to the $k$-th condition. We can further obtain a one-to-one mapping based on $T$ under assumptions~\cite{Courty2017Optimal}. Fig.~\ref{fig:evaluation} shows a comparison between our proposed and previous criteria.

\noindent{\bf Discussions.}  
Based on our criterion, a WS-CSL model gets high accuracy only when all conditions can be explained with certain learned embeddings, and the model predicts triplets from diverse aspects well with different embeddings. 
Thus our criterion depicts the ``semantic coverage'' of $\Psi_K$, which also reveals the gap between $\Psi_K$ in WS-CSL and its supervised counterpart. 
We may compute the cost $C$ over the validation data with a small amount of ground-truth condition labels, and use the obtained alignment to compute the final accuracy on the test set.
Benefiting from this condition-embedding alignment, our criterion naturally avoids the issue from the reversed triplets shown in Fig.~\ref{fig:teaser}. 

\section{{\textbf{\scshape DiscoverNet}} for Weakly-Supervised CSL}\label{sec:method}
\begin{figure}[!t]
	\vspace{-2mm}
	\centering
	\begin{minipage}[h]{\linewidth}
		\centering
		\includegraphics[width=\linewidth]{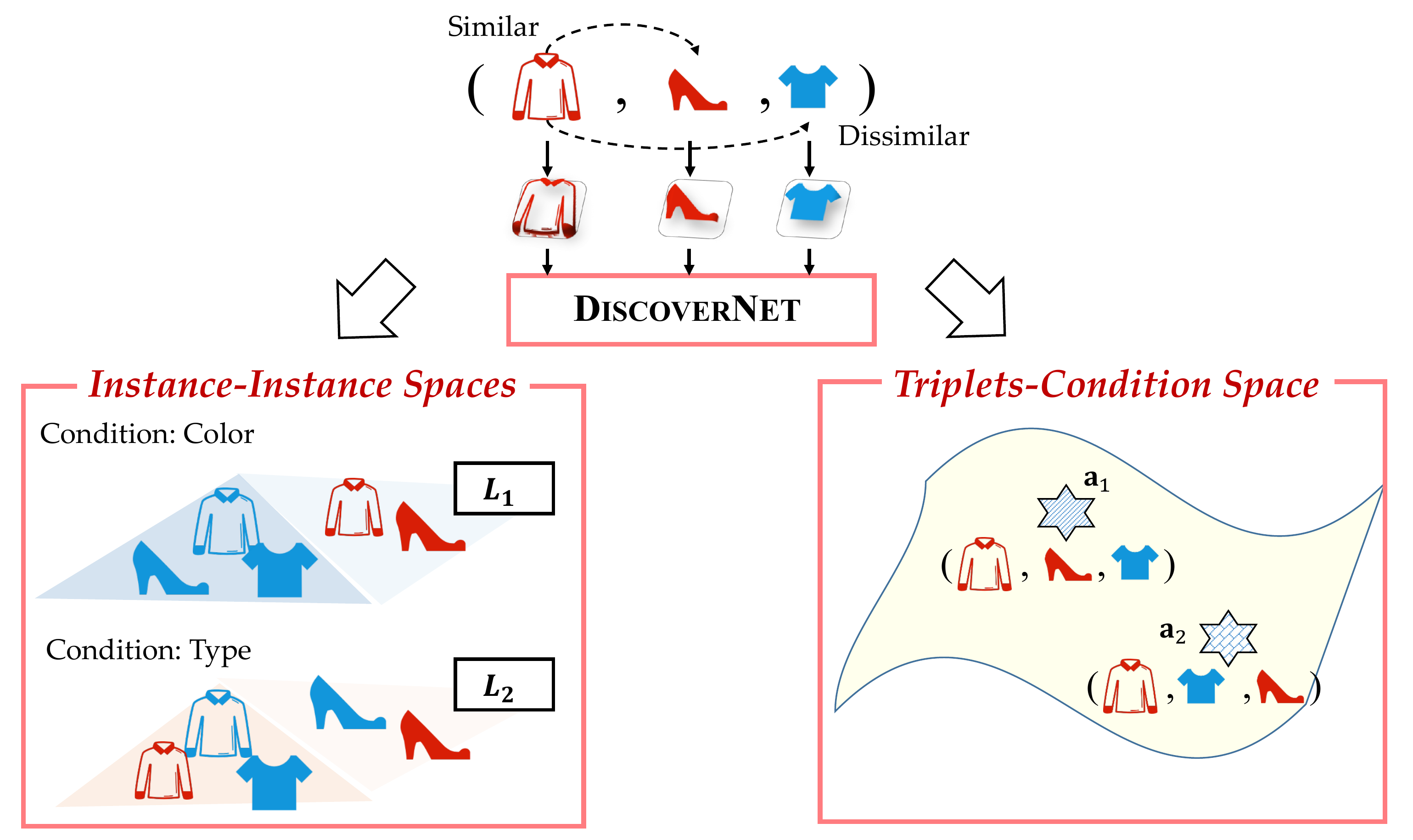}\\
	\end{minipage}
	\vspace{-2mm}
	\caption{{Illustration of {\name} with two types of space. The embedding space could be decomposed based on two different projections ($L_{1}, L_{2}$) and each of them has their own ``instance-instance'' linkage preference. For example, the first similarity condition is about ``Color'' and the second one focuses on ``Type''. Besides, we identify the latent similarity condition through the ``triplets-condition'' space, where a triplet is summarized as a vector by $g(\cdot)$ and we match this triplet with all condition anchors $(a_{1}, a_{2})$.}
	The ability of {\name} to learn decomposed embeddings are demonstrated in Section~\ref{sec:ablation}.
	}\label{fig:workflow}
	\vspace{-5mm}
\end{figure}

We'd like to learn the embedding set $\Psi_K$ to cover all rich semantics, and the comparison relationship under each target condition could be revealed by a certain conditional embedding $\psi_k$. 
We propose Distance Induced Semantic Condition VERification Network~(\name) for WS-CSL (illustrated in Fig.~\ref{fig:workflow}). {\name} introduces a ``triplets-condition'' space to match condition scores of a triplet, which automatically selects or fuses multiple pairwise distances measured by ``instance-instance'' spaces.
As we discussed before, a WS-CSL model could violate the semantic constraints over artificially reversed triplets, which influences the coverage of semantics. 
We consider a set module to avoid those cases during training, which maps various triplets with the same set of instances to one condition. Furthermore, we add a regularizer to force the model to select diverse conditions if both original and reversed triplets are predicted as valid ones.

\subsection{Decompose and Fuse Hierarchical Spaces}
We assume condition-specific embeddings $\Psi$ have different projections $\mathcal{L}_K$ but share the same $\phi$.\footnote{Since conditions are not independent, it is not necessary to match each condition with an embedding $\psi_k$. More discussions are in experiments.} Then we have
{ \begin{equation} \label{eq:diffk}
\left\{\mathbf{Diff}^k_\tau =\mathbf{Dis}_{L_k}^2\left(\phi(\x),\phi(\z)\right)-\mathbf{Dis}_{L_k}^2\left(\phi(\x),\phi(\y)\right)\right\}_{k = 1}^K.
\end{equation}}The space measured by projection $L_k$ biases towards a ``local'' view of the embedding $\phi$, and allows inconsistent comparisons across different ``instance-instance'' spaces.

The overall linkages between objects are measured based on a fused distance over condition-specific comparisons. Different from the indicator in Eq.~\ref{eq:csl_obj} selecting the distance, we use a multinomial distributed variable $c_\tau\in\{0,1\}^K$ to denote the latent condition label of a triplet $\tau$.
There are $K$ binary elements in $c_\tau$.
We optimize $\phi$ and $\mathcal{L}_K$ jointly:
\begin{equation}
\min_{\phi, \mathcal{L}_K} \;\sum_{\tau\in\mathcal{T}}\; \ell\left(\mathbb{E}_{c_\tau}\left[\mathbf{Diff}^k_\tau\right] - \gamma\right)\;.\label{eq:obj}
\end{equation}
The condition of the triplet is revealed by the {\em expected distance} w.r.t. the distribution of $c_\tau$~\cite{Ye2017Learning}, which {\em selects or fuses} the related conditions of the triplet.
If a triplet could only be valid under the $k$-th condition, we expect $c_\tau$ has the $k$-th element $c_\tau^k=1$ and other elements equal 0. Then the expected distance activates the $k$-th metric, and the target neighbor (resp. impostor) is pulled close (resp. pushed away) with $L_k$. The $k$-th embedding becomes more specific for the condition. If the triplet is related to more conditions, we expect $\tau$ to fuse the semantics from multiple distances together.
Therefore, values in $\tau$ determine whether to select or fuse conditional embeddings, which emphasizes the semantic coverage or comparison prediction, respectively.

Given $\mathbb{E}_{c_\tau}[\textbf{Diff}_\tau^k] = \sum_{k=1}^K\Pr(c_\tau^k=1)\textbf{Diff}_\tau^k$, the variable $c_\tau$ indicates the influence of the posterior probability $\Pr(c_\tau^k=1)$ that a triplet belongs to the $k$-th condition. We introduce a ``triplets-condition'' space, where a triplet is embedded to a point with mapping $g(\cdot)$.
$g$ summarizes the triplet $\tau$ and maps the set of embeddings in $\tau$ into a $d$-dimensional vector. We expect triplets with similar conditions will be close. 
Furthermore, $K$ learnable anchors $\{\aaa_k\in\mathbb{R}^d\}_{k=1}^K$ captures those similarity conditions, and we match a triplet to its closest anchor:
\begin{align}
\Pr(c_\tau^k=1) &\;\sim\; \mathbf{Sim}(g(\tau), \;\aaa_k)\notag\\
&\;=\; \frac{\exp\left(\cos(g(\tau),\; \aaa_{k})/\varsigma\right)}{\sum_{k'}^K \exp\left(\cos\left(g(\tau),\; \aaa_{k'}\right)/\varsigma\right)}\;.\label{eq:cosine}
\end{align}
We implement the similarity with cosine. $\varsigma>0$ is the temperature. The larger the $\varsigma$, the more uniform $\Pr(c_\tau^k=1)$ is.
If a triplet is related to the $k$-th condition, then its transformed vector $g(\tau)$ is close to the anchor $\aaa_k$. Then a larger $\mathbf{Sim}(g(\tau), \;\aaa_k)$ emphasizes the $k$-th condition when computing the expected distance. 
Benefited from the ``decompose and fuse'' manner in Eq.~\ref{eq:cosine}, a triplet that has a clear condition tendency will be close to a particular anchor, which makes $c_\tau$ becomes one-hot, and the embedding is highly correlated with the corresponding condition. Then embeddings will cover more semantics. 
Those ambiguous triplets will have a nearly uniform $c_\tau$. Although hard to differentiate semantics in this case, their fused distances take advantage of all embeddings to predict the validness of triplets and improve the comparison prediction ability.

\noindent{\bf Instance Space for Semantic Discovery.}
Given $L_k\in\mathbb{R}^{d\times d}$, we define the $k$-th conditional embedding $\psi_k(\x)$ as
{
\begin{equation}
\psi_k(\x) = \phi(\x) (I + L_k) = \phi(\x) + \phi(\x)L_k\;.
\end{equation}}We obtain the local embedding $\psi_k$ in a residual form, where $L_k$ encodes the {\em similarity bias} of the $k$-th condition based on the general embedding $\phi$. If $\phi$ is discriminative enough for a particular similarity condition, we do not need to over-allocate the local metric. In other words, strong $\phi$ makes $L_k$ degenerate to zero. In summary, distances for different similarity conditions in Eq.~\ref{eq:diffk} are calculated based on $\psi_k$.

\begin{figure}[!t]
	\centering
	\begin{minipage}[h]{\linewidth}
		\centering
		\includegraphics[width=\linewidth]{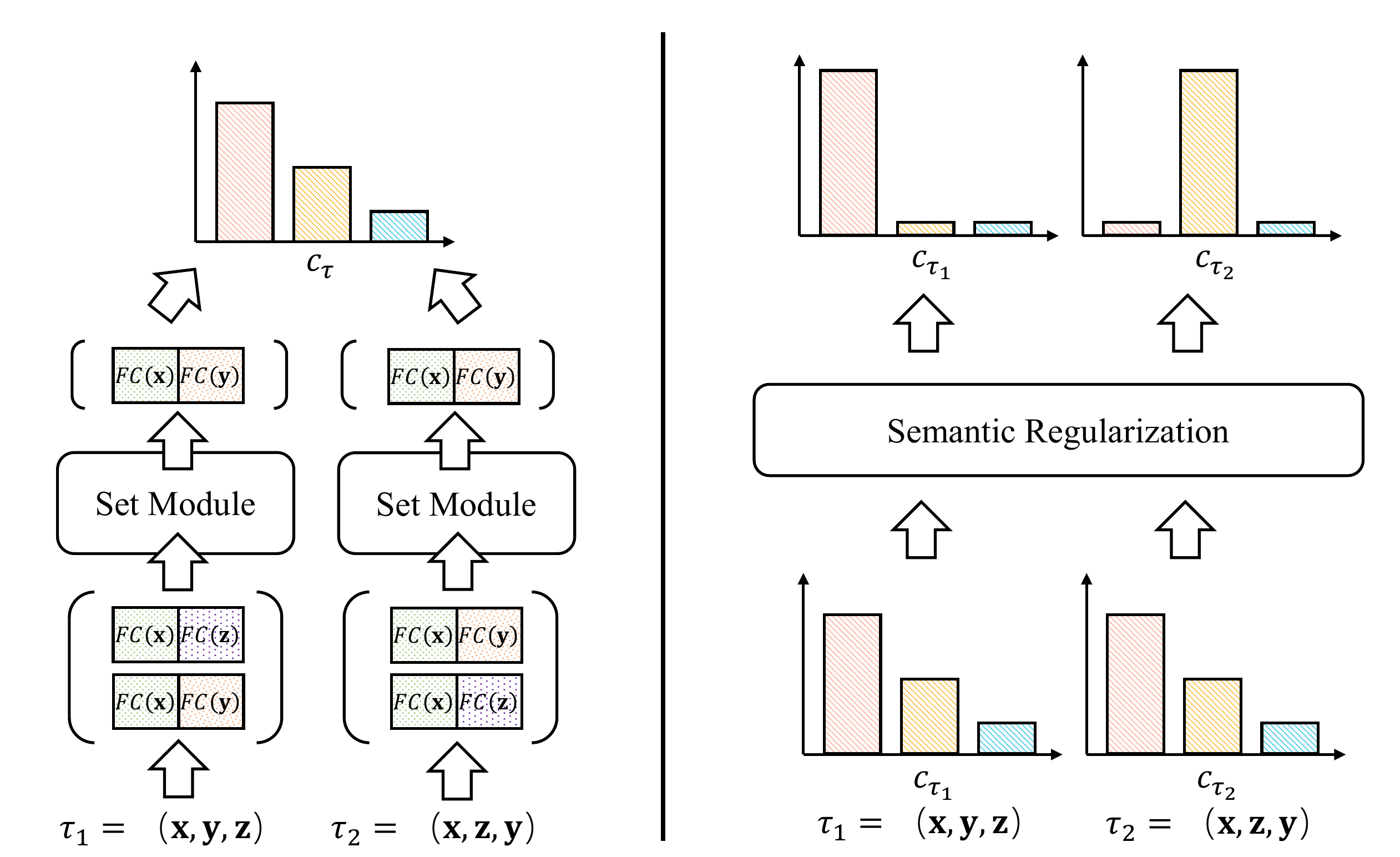}\\
	\end{minipage}
	\vspace{-1mm}
	\caption{Illustration of the set module (left) and semantic regularization (right) to deal with artificially reversed triplets. 
	}\label{fig:workflow2}
	\vspace{-5mm}
\end{figure}

\subsection{Condition Space for Semantic Fusion}\label{sec:triplet_space}
Based on our analyses in Section~\ref{sec:evaluation}, a WS-CSL model could violate the semantic constraints in vanilla training. A {\em too flexible} mapping $g$ will make the model treat both a triplet and its reversed version as correct ones while activating almost the same conditional embedding $\psi_k$. In other words, the model may focus on fusing those conditions with $g$ and collapse all semantics into one conditional space.

Since semantic constraints on a WS-CSL model are limited, we address the issue with the help of {\em artificially reversed triplets}.
For a pair of original triplet and its reversed version, we restrict the model's flexibility by either weakening the representation ability of $g$ with a set module or adding another semantic regularizer (illustrated in Fig.~\ref{fig:workflow2}).

\noindent{\bf Set Module.}
A natural way to implement $g$ is to capture the order of instances in a triplet with a sequential model~\cite{Hochreiter1997Long,Tan2019Learning
}.
As we mentioned, a triplet $\tau_1 = (\x, \y, \z)$ and its reversed version $\tau_2=(\x,\z,\y)$ should belong to different conditions.
However, without any constraints, we find a sequential module has the ability to weight conditions in different manners for $\tau_1$ and $\tau_2$, but it is likely to work in a ``lazy'' way. 
In detail, a model may learn a strong conditional embedding $\psi_k$ and map both $\tau_1$ and $\tau_2$ to similar distributions that have larger weights on $\psi_k$ and small weights on others.
Then, rather than making learned embeddings associated with corresponding conditional semantic meanings, a model tends to improve the fusion module and use partial embeddings to cover all semantics.

Therefore, we consider a set module to make $g$ agnostic with the order in the triplet. 
In particular, we use the pairwise concatenation of instance embeddings as the input of $g$, \ie, for a triplet $\tau_1$, we obtain its embeddings, $\phi(\tau) = (\phi(\x), \phi(\y), \phi(\z))$ and get the augmented set: 
\begin{equation*}
\mathcal{P} = \{[\phi(\x), \phi(\y)], [\phi(\x), \phi(\z)]\}\;,
\end{equation*}
which encodes the pairs in $\tau_1$. Since $[\phi(\y), \phi(\z)]$ is not included in $\mathcal{P}$, we get the same representation for $\tau_1$ and the reversed $\tau_2$.  
Moreover, to take a holistic view of $\tau$, we transform with element-wise maximization~\cite{Zaheer2017Deepsets}:
\begin{equation}
g(\tau) = \mathbf{FC}\; \max_{\p\in\mathcal{P}} \;\mathbf{FC}(\p)\;.\label{eq:activation}
\end{equation}
$\mathbf{FC}$ is a fully connected network with one hidden layer and ReLU activation, which projects the given input to a same-dimensional output. With Eq.~\ref{eq:activation}, the pairwise relationship in the triplet is evaluated first, and the most evident one is activated to represent the whole triplet. In summary, the set module restricts the ability of $g$ and makes one triplet and its reversed version have the same representation. 
Although it is a bit counter-intuitive, the special configuration of $g$ avoids the ambiguous update directions of the model so that the conditional embeddings should be more discriminative to cover the target semantics. We also investigate the Transformer~\cite{Vaswani2017Attention} implementation of $g$ in the supplementary.

\noindent{\bf Semantic Regularizer.} We can also think from another aspect by regularizing a representative $g$.
We set Eq.~\ref{eq:activation} to \begin{equation*}
\mathcal{P} = \{[\phi(\x), \phi(\y)], [\phi(\x), \phi(\z)], [\phi(\y), \phi(\z)]\}\;,\label{eq:activation2}
\end{equation*}
covering the sequential information of instances.
Due to an additional pair in $\mathcal{P}$, different orders of instances in $\tau_{1}$ and $\tau_{2}$ do not share the same output. 
In this case, it is abnormal when the model predicts $\tau_{1}$ and $\tau_{2}$ based on the same condition $k$. 
We construct a semantic regularization to avoid this case. If $\mathbf{Diff}^k_{\tau_1} > 0$ and $\mathbf{Diff}^k_{\tau_2} > 0$ (the model treats two triplets as valid ones),\footnote{We check this condition with the current model during training. Parameters in $\mathbf{Diff}^k_{\tau}$ here are detached without gradient back propagation.} we add the following semantic regularizer to make the activated conditions diverse:
\begin{equation}
{\rm Reg}(g) = \lambda \;{\sum_{k=1}^K\min\;(\mathbf{Sim}(g(\tau_{1}), \;\aaa_k),\; \mathbf{Sim}(g(\tau_{2}), \;\aaa_k))}
\;.\label{eq:reg}
\end{equation}
In other words, if $\tau_1$ and $\tau_2$ are predicted as valid ones, we minimize the similarity between their condition distributions $c_{\tau_1}$ and $c_{\tau_2}$. The similarity between these two multinomial distributions are measured with Histogram Intersection Kernel (HIK)~\cite{Grauman2005The,Wu2010Fast,Wu2013C4}, which is the sum of element-wise minimum of two distributions. 
Therefore, with the help of Eq.~\ref{eq:reg}, we explicitly enforce the WS-CSL model considers different conditional embeddings to explain triplets, which further improve the semantic coverage of $\Psi_K$.

\noindent{\bf Discussions.} There are two implementations of {\name}, using the set module or the semantic regularizer. The two strategies are designed from different aspects to make the WS-CSL model contain rich semantics as much as the supervised model. Since the set module avoids the diverse selection for reversed triplet naturally, it satisfies the regularizer directly. Thus, it does not help if we combine two strategies together.

\section{Experiments}\label{sec:exp}
We verify the effectiveness of {\name} 
over benchmarks based on our new criterion. 
Ablation studies and visualization results demonstrate {\name} learns conditions successfully as the supervised methods.
Detailed setups and more results are in the supplementary.
 
\subsection{Experimental Setups}
\noindent\textbf{Datasets.} 
\textbf{UT-Zappos-50k Shoes} contains 50,025 images of shoes with four similarity conditions collected online~\cite{Yu2014finegrained,Yu2017Semantic}. 
Following~\cite{Veit2017Conditional,Tan2019Learning,Nigam2019Towards}, we discretize the ``height'' condition and resize all images to 112 by 112. There are 200,000, 20,000, and 40,000 triplets following splits in~\cite{Veit2017Conditional} for training, validation and test, respectively.
\textbf{Celeb-A Faces} has 202,599 face images of different identities~\cite{Liu2015Deep}. 
8 of the 40 attributes (conditions) are selected for analysis~\cite{Nigam2019Towards}.
We resize all images to 112 by 112. 400,000/80,000/160,000 triplets are used for model training/validation/test.
We construct a more difficult \textbf{Celeb-A$^\dagger$} by combining related binary attributes in Celeb-A together, 
where each multi-choice condition has 5-7 discrete values,
We apply the same configuration for Celeb-A variants.

\noindent\textbf{Splits.}
All models are trained and evaluated over triplets in~\cite{Veit2017Conditional}. Since there are no published triplets over Celeb-A, we randomly sample triplets by ourselves. Equal number of triplets for each attribute are sampled from the standard training, validation, and test split of Celeb-A~\cite{Liu2015Deep}. 
For each triplet, we organize instances with the same attribute label into a similar pair. Otherwise, we think they are dissimilar. 

\noindent\textbf{Criterion.} We utilize our proposed new criterion to evaluate WS-CSL methods. After constructing a mapping from condition to learned embeddings with greedy or OT strategies, we compute the average triplet prediction accuracy as in the supervised case. We denote the results with two strategies as GR Accuracy and OT Accuracy, respectively.

\noindent\textbf{Comparison Methods.}
We compare {\name} with both supervised method Conditional Similarity Networks~(CSN)~\cite{Veit2017Conditional} and two WS-CSL methods, \ie, Latent Similarity Networks~(LSN)~\cite{Nigam2019Towards} and Similarity Condition Embedding Network~(SCE-Net)~\cite{Tan2019Learning}.

\noindent\textbf{Implementation Details.}
Following~\cite{Veit2017Conditional,Nigam2019Towards,Tan2019Learning}, we use ResNet-18~\cite{He2016Residual} to implement $\phi$. Different from previous literature fine-tuning the backbone based on the weights pre-trained on ImageNet~\cite{Deng2009Imagenet}, we also consider the case that we train the full model from scratch. 
The model with the best accuracy over the validation set is selected for the final test.

\subsection{Benchmark Evaluations}
\begin{table}[tbp]
	\centering
	\caption{GR accuracy and OT accuracy on UT-Zappos-50k. We investigate two cases that training the model from scratch and fine-tune the model with pre-trained weights. CSN is the fully supervised CSL method which utilizes condition labels during training. We make the best WS-CSL results in each case in bold.}
	\vspace{-3mm}
	\tabcolsep 2pt
	\begin{tabular}{c|cc||cc}
		\addlinespace
		\toprule
		Setups $\rightarrow$ & \multicolumn{2}{c||}{w/ pretrain} & \multicolumn{2}{c}{w/o pretrain}\\
		\midrule
		Criteria $\rightarrow$ & GR Acc. & OT Acc.& GR Acc. & OT Acc. \\
		\toprule
		CSN~\cite{Veit2017Conditional}  & 87.86 & 87.86 & 82.14 & 82.14  \\
		LSN~\cite{Nigam2019Towards} & 76.26 & 75.90 & 71.49 & 68.49 \\ 
		SCE-Net~\cite{Tan2019Learning}	 & 72.21 & 71.15 & 64.08 & 61.03\\
		\midrule
		{\name}$_{\rm Set}$ & 76.98 & 75.68 & {\bf 74.67} & {\bf 74.13}\\
		{\name}$_{\rm Reg}$ & {\bf 77.84} & {\bf 77.68} & 72.99 & 71.46\\
		\bottomrule
	\end{tabular}\label{table:zappos}
\vspace{-3mm}
\end{table}
\noindent{\bf UT-Zappos-50k.} The results of {\name} and comparison methods over UT-Zappos-50k are listed in Table~\ref{table:zappos}, which contains the greedy accuracy and OT accuracy over both pre-trained and non-pre-trained weights. 
We re-implement all comparison methods. {\name}$_{\rm Set}$ and {\name}$_{\rm Reg}$ denote the variant using set module and semantic regularization, respectively. 

By observing both setups and criteria, CSN becomes the ``upper-bound'' for WS-CSL methods. The main reason is that ground-truth condition labels in CSN associate an embedding with a particular condition during training. 
The supervised CSN gets the same greedy and OT accuracy. The explicit supervision in CSN makes those embeddings biased towards different semantics, which gets the same one-to-one OT mapping with the greedy strategy. 
WS-CSL methods get lower OT accuracy than the corresponding greedy accuracy. The main reason is that OT accuracy requires a one-to-one mapping between embeddings and conditions, where those conditions would be related. While greedy accuracy allows one embedding to handle multiple conditions. 

As we discussed before, SCE-Net tends to fuse the semantic meaning of embeddings with its self-attention module, so each of its learned embeddings is hard to cover a specific condition. By contrast, LSN performs better in our criteria, benefiting from its multi-choice learning paradigm. Our {\name} can get the best performance among WS-CSL methods. In detail, {\name}$_{\rm Set}$ works better when training from scratch and {\name}$_{\rm Reg}$ performs well with the pre-trained weights. One possible reason is that the pre-trained weights are strong and make the model (especially the mapping function $g$) too flexible, so an explicit regularization helps more. In summary, our criterion reveals how much similar a WS-CSL model performs to a supervised one.
\begin{figure*}
	\centering
	\vspace{-2mm}
	\begin{minipage}[h]{0.24\linewidth}
		\centering 
		\includegraphics[width=\textwidth]{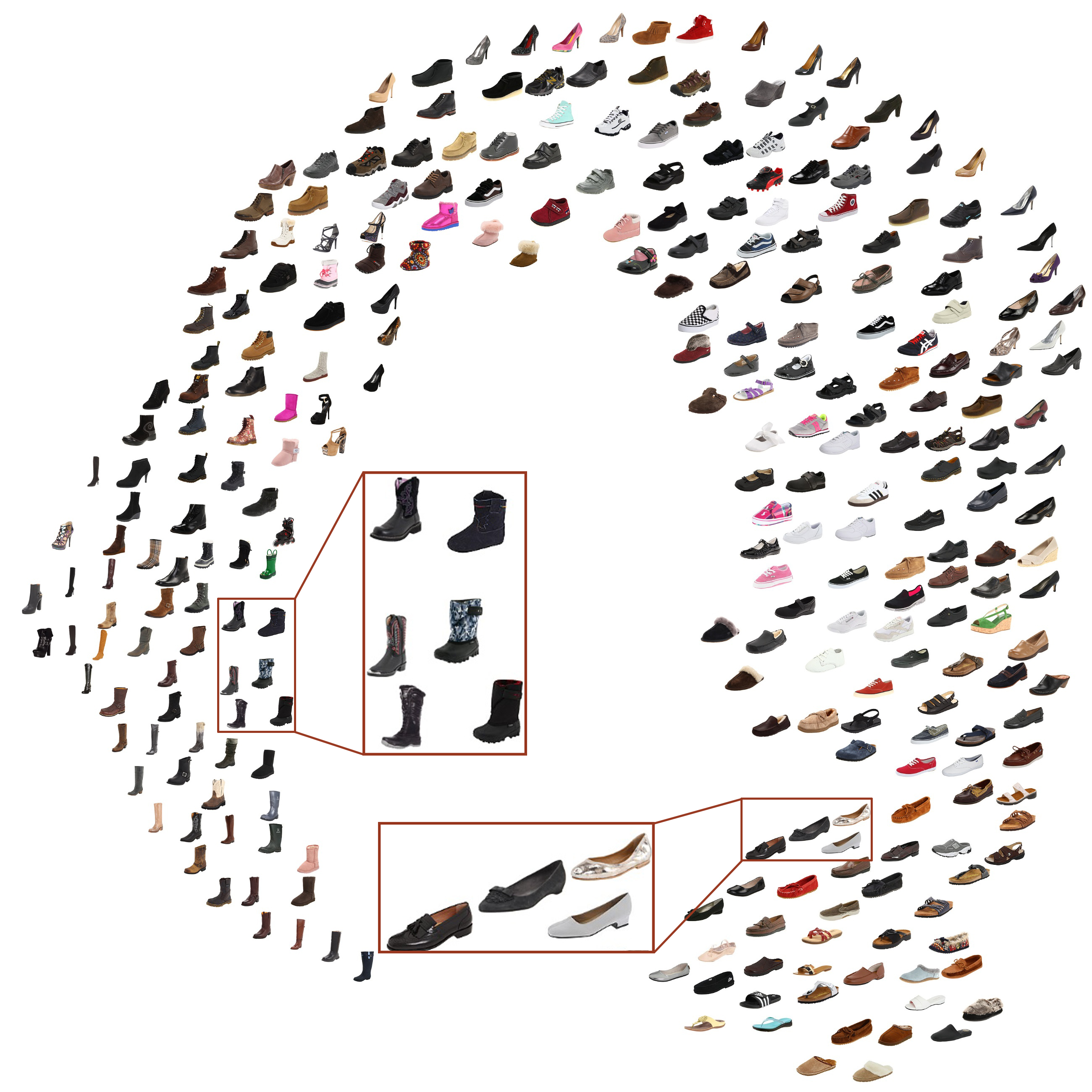}\\
		\mbox{({\it a}) {Functional Types}}
	\end{minipage}
	\begin{minipage}[h]{0.24\linewidth}
		\centering 
		\includegraphics[width=\textwidth]{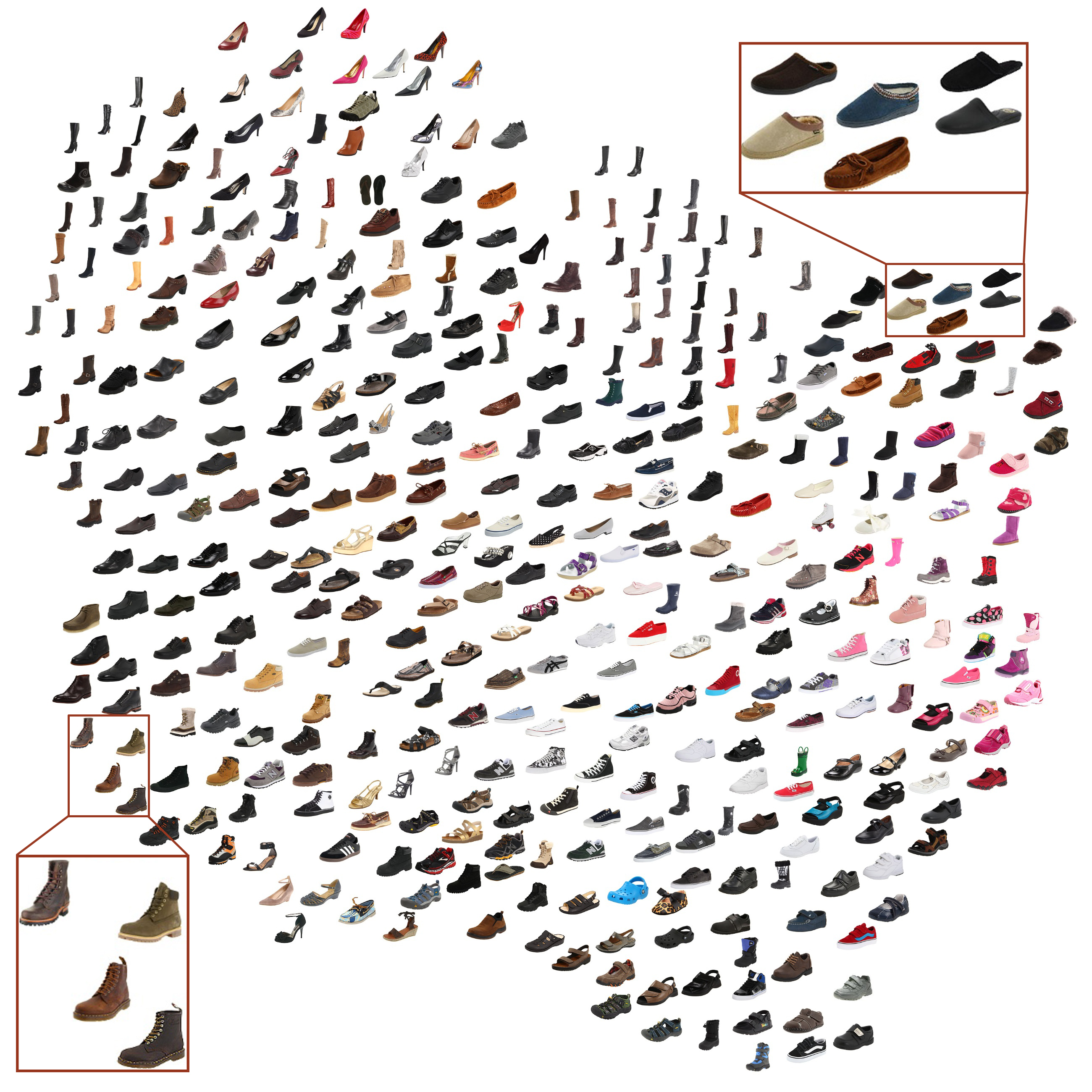}\\
		\mbox{({\it b}) {Closing Mechanism}}
	\end{minipage}
	\begin{minipage}[h]{0.24\linewidth}
		\centering 
		\includegraphics[width=\textwidth]{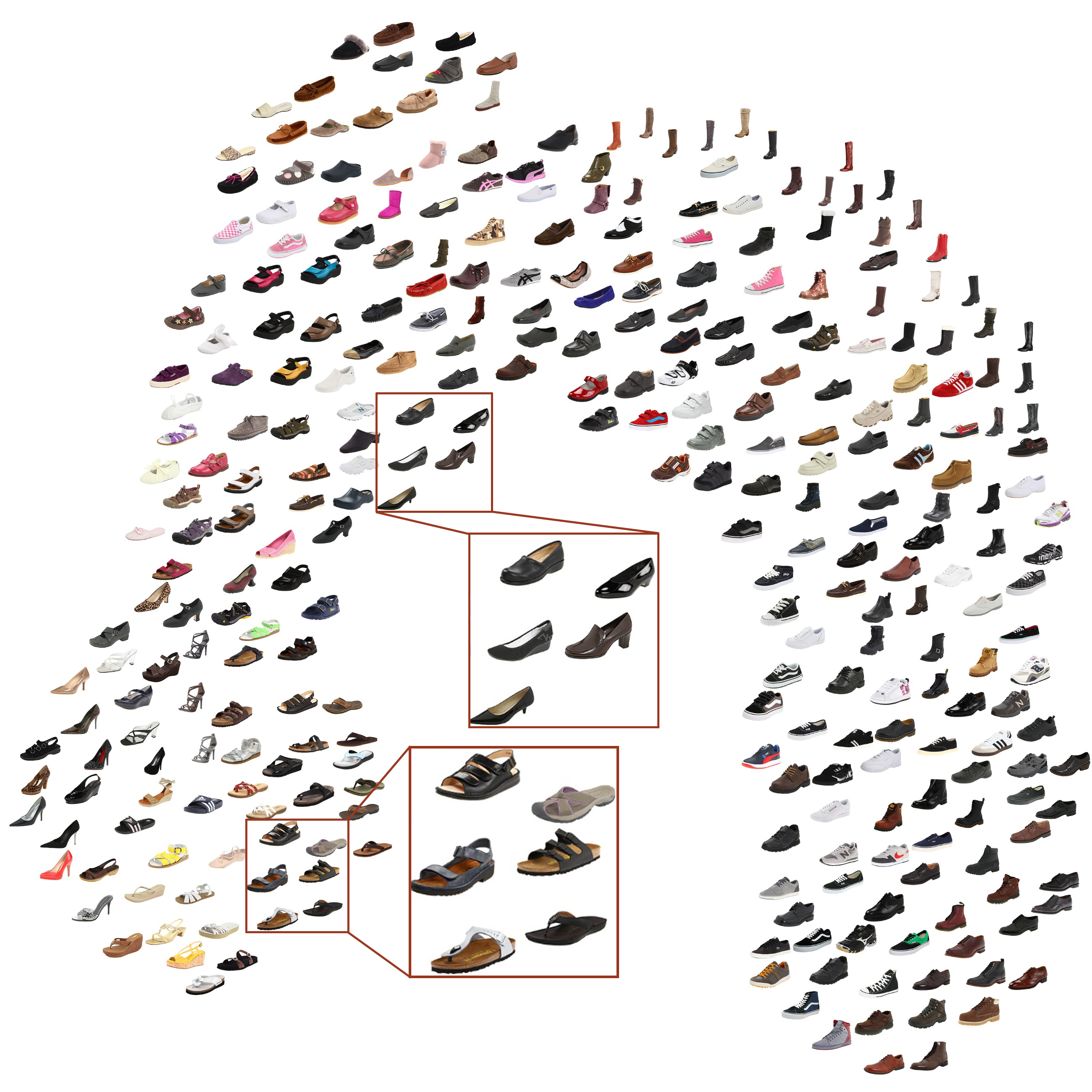}\\
		\mbox{({\it c}) {Suggested Gender}}
	\end{minipage}
	\begin{minipage}[h]{0.24\linewidth}
		\centering 
		\includegraphics[width=\textwidth]{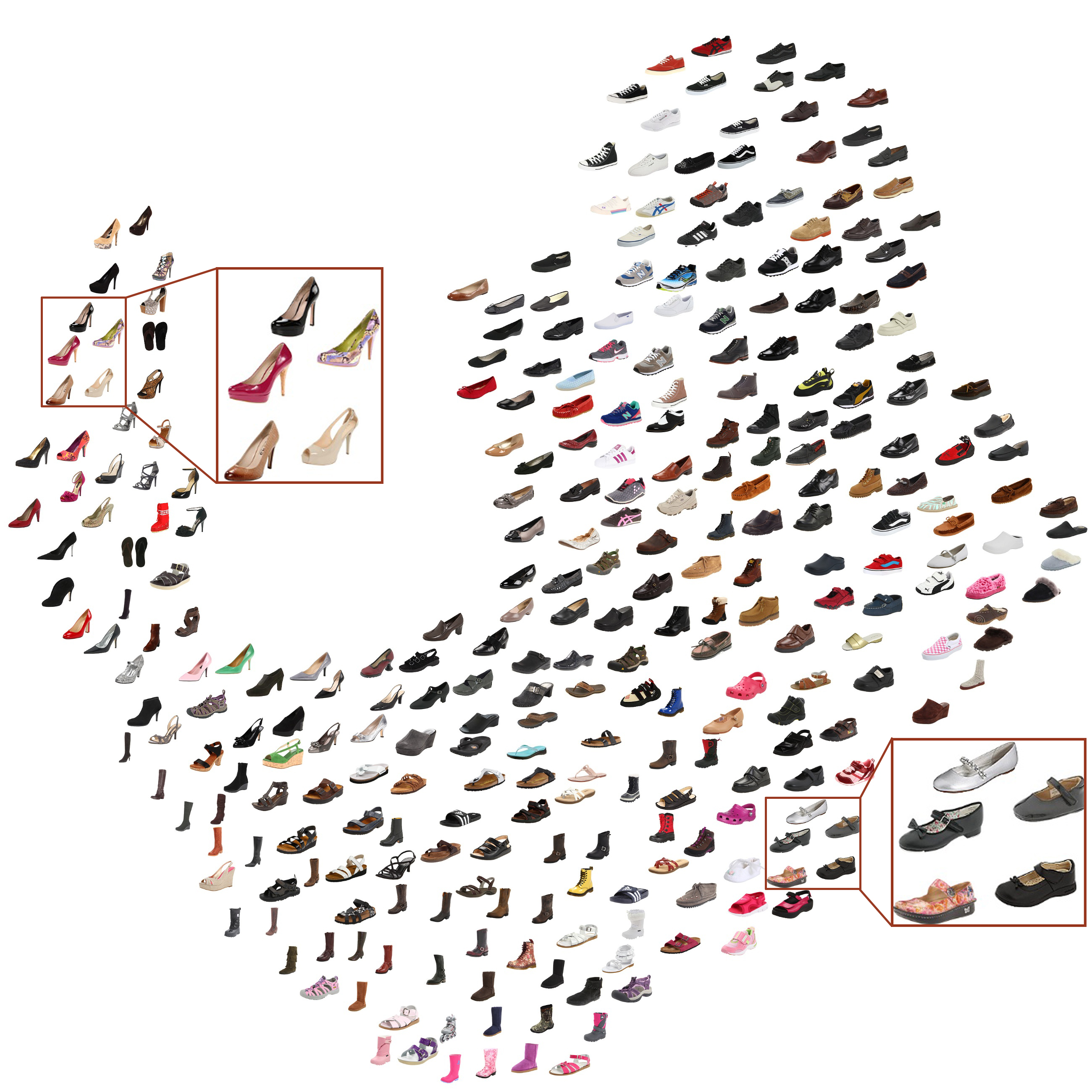}\\
		\mbox{({\it d}) {Height of Heels}}
	\end{minipage}
	\caption{TSNE of the learned embeddings for each of the four conditions (\ie, functional types, closing mechanism, suggested gender and height of heels) on UT-Zappos-50k dataset based on {\name}$_{\rm Reg}$.}\label{fig:TSNE}
	\vspace{-2mm}
\end{figure*}

\begin{table}[tbp]
	\centering
	\caption{Greedy accuracy and OT accuracy on 8-condition Celeb-A (binary conditions) and its attribute merged variant Celeb-A$^\dagger$ with five multi-choice conditions, respectively. All methods are trained from scratch. More results are in the supplementary.} 
	\tabcolsep 2pt
	\vspace{-3mm}
	\begin{tabular}{c|cc||cc}
		\addlinespace
		\toprule
		& \multicolumn{2}{c||}{Celeb-A} & \multicolumn{2}{c}{Celeb-A$^\dagger$}\\
		\midrule
		Criteria $\rightarrow$ & GR Acc. & OT Acc. & GR Acc. & OT Acc. \\
		\midrule
		CSN~\cite{Veit2017Conditional}  & 83.41 & 83.41 & 67.72 & 67.72 \\
		LSN~\cite{Nigam2019Towards} & 70.29 & 69.89 & 57.02 & 56.33\\
		SCE-Net~\cite{Tan2019Learning}	 & 69.47 & 51.81 & 51.49 & 50.64\\
		\midrule
		{\name}$_{\rm Set}$ & {\bf 78.45} & {\bf 77.98} & {\bf 64.57} & {\bf 63.98}\\ 
		{\name}$_{\rm Reg}$ & 78.04 & 76.96 & 57.43 & 56.94\\
		\bottomrule
	\end{tabular}\label{table:celebra}
	\vspace{-3mm}
\end{table}

\noindent{\bf Celeb-A.} In Table~\ref{table:celebra}, we investigate (the 8 attribute) Celeb-A and its variant Celeb-A$^\dagger$ with model trained from scratch.
{\name}$_{\rm Set}$ and {\name}$_{\rm Reg}$ still get better performance than other WS-CSL counterparts, while CSN is still the `upper bound' due to the help of condition labels. 

\subsection{Ablation Studies and Visualizations}\label{sec:ablation}
We analyze the properties of {\name} and show the visualization results.
More analysis such as the help of the WS-CSL embeddings given limited conditional supervisions are in the supplementary. 

\begin{table}[tbp]
	\centering
	\caption{Influence of the embedding number (the number of projections in $\mathcal{L}_K$) for {\name}$_{\rm Set}$ and {\name}$_{\rm Reg}$ on Celeb-A. Models are fine-tuned with pre-trained weights.} 
	\tabcolsep 3pt
	\vspace{-3mm}
	\begin{tabular}{c|cc||cc}
		\addlinespace
		\toprule
		& \multicolumn{2}{c||}{{\name}$_{\rm Set}$} & \multicolumn{2}{c}{{\name}$_{\rm Reg}$}\\
		\midrule
		\# Projections & GR Acc. & OT Acc. & GR Acc. & OT Acc. \\
		\midrule
		2  & 78.28 & 77.74 & 71.92 & 71.82 \\
		4  & 79.58 & 77.60 & 73.19 & 71.32\\
		6  & 80.41 & 77.69 & 75.53 & 73.87\\
		8  & 80.65 & 78.81 & 75.19 & 74.07\\ 
		10 & 81.35 & 78.68 & 76.09 & 73.77\\
		\bottomrule
	\end{tabular}\label{tab:nconditions}
	\vspace{-3mm}
\end{table}

\begin{figure}
	\centering
	\begin{minipage}[h]{0.95\linewidth}
		\centering 
		\includegraphics[width=\textwidth]{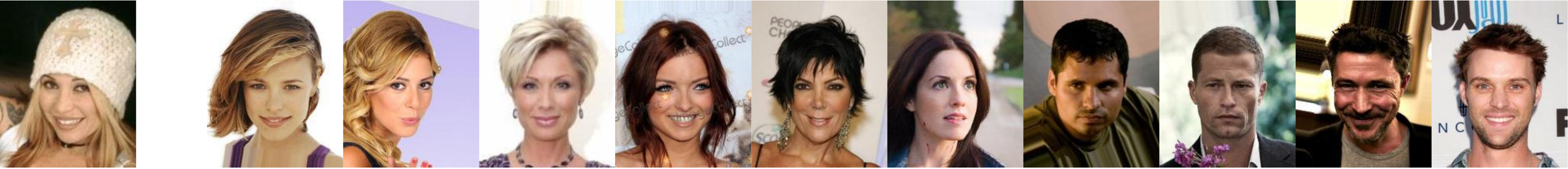}\\
		\mbox{({\it a}) {5 o Clock Shadow}}
	\end{minipage}
	\begin{minipage}[h]{0.95\linewidth}
		\centering 
		\includegraphics[width=\textwidth]{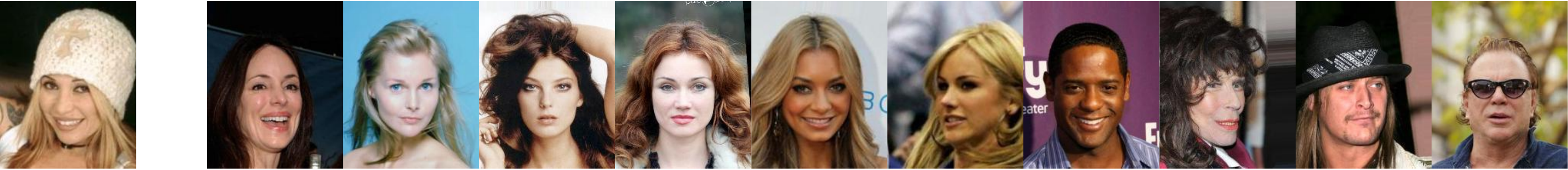}\\
		\mbox{({\it b}) {Eyeglasses}}
	\end{minipage}
	\begin{minipage}[h]{0.95\linewidth}
		\centering 
		\includegraphics[width=\textwidth]{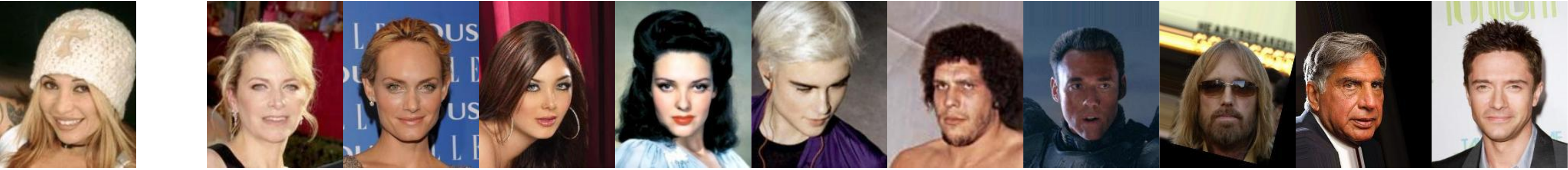}\\
		\mbox{({\it c}) {Male}}
	\end{minipage}
	\begin{minipage}[h]{0.95\linewidth}
		\centering 
		\includegraphics[width=\textwidth]{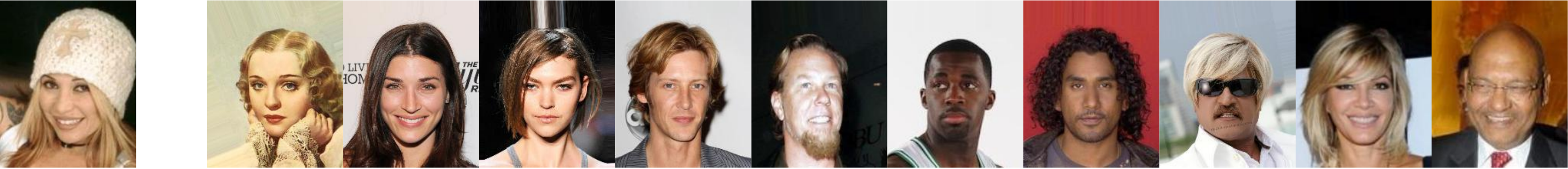}\\
		\mbox{({\it d}) {Wearing Lipstick}}
	\end{minipage}
	\caption{Visualization of the image retrieval results for each of the four conditions on Celeb-A dataset with the learned embedding of {\name}$_{\rm Reg}$. The first image in each row is the anchor, and faces are ranked by distances to the anchor in an ascending order.}\label{fig:retrieval}
	\vspace{-5mm}
\end{figure}
\noindent\textbf{Influence of condition number configuration.}
We set the number of embeddings in benchmarks the same as the number of ground-truth conditions, which is 8 in Celeb-A. 
We show the change of the two criteria along with the increase of the embedding number in Table~\ref{tab:nconditions}. 
Both {\name}$_{\rm Set}$ and {\name}$_{\rm Reg}$ get higher OT accuracy when the number of embeddings increases from two to eight, and decreases from eight to ten. We conjecture that allocating too many local metrics interfere with each other and their related semantics partially overlap. Besides, greedy accuracy shows a generally incremental trend since it allows one embedding to handle multiple conditions.

\noindent\textbf{Visualization of semantic embeddings.}
To better illustrate how our {\name} learns for different semantics, we provide TSNE visualizations for each of the learned semantic spaces $\{\psi_k\}_{k=1}^K$ on UT-Zappos-50k in Fig.~\ref{fig:TSNE}. {\name} captures the variety of conditions and learns different embeddings for the dataset with good interpretability. Typically, on the condition ``height of heels'', the heel height of the shoes is decreasing from the left-top of the embedding space to the bottom, then to the right-top. 

\noindent\textbf{Visualization of conditional image retrieval.}
We provide image-retrieval visualizations for four conditions on Celeb-A in Fig.~\ref{fig:retrieval}.
We keep the anchor image, and retrieve its neighbor from a set of randomly collected candidates with different local embeddings. Benefiting from the correspondence obtained when computing the OT accuracy, we can qualitatively measure whether a local embedding could reveal the corresponding semantic via its ranking of images. 
For example, on the ``Male'' condition, since the anchor image has the label ``Female'', {\name} makes all images labeled ``Female'' close while pushing images related to ``Male'' away. The results indicate our {\name} can cover latent semantics of data and each of its learned conditional embedding $\psi_k$ corresponds to a particular meaningful semantic. 

\section{Conclusion}
We revisit WS-CSL and observe that evaluating the quality of the model without specifying concrete conditions produces biased accuracy.
Thus, we match multiple learned embeddings with ground-truth conditions in advance before predicting the correctness of given triplets, which simultaneously considers the ability of triplet prediction and semantic coverage.
We also utilize a set module or a semantic regularizer in our proposed {\name} to emphasize the correspondence between a conditional embedding and a semantic condition.
{\name} outperforms other WS-CSL methods on benchmarks with different criteria. 

\noindent{\bf Limitations.} Our new criterion evaluates WS-CSL in a ``supervised'' manner by assigning multiple learned embeddings to target conditions. 
The criterion does not fit the case when the goal is not to learn a model similar to its supervised counterpart, \eg, to distinguish whether triplets are correct and explain their validness as much as possible.

\noindent{\bf Acknowledgments.}
This research was supported by National Key
R\&D Program of China (2020AAA0109401), NSFC (62006112, 61921006, 62176117),  Collaborative Innovation Center of Novel Software Technology and Industrialization, NSF of Jiangsu Province (BK20200313).

\newpage

{\small
\bibliographystyle{ieee_fullname}
\bibliography{Discover}
}

\end{document}


\title{Supplemental Material of\\Identifying Ambiguous Similarity Conditions via Semantic Matching}

\author{Han-Jia Ye\qquad Yi Shi\qquad De-Chuan Zhan \\
State Key Laboratory for Novel Software Technology, Nanjing University
\\
{\tt\small \{yehj, shiy, zhandc\}@lamda.nju.edu.cn}}

\maketitle

\begin{abstract}
There are four parts in this supplementary:
\begin{itemize}
	\item The details to reproduce Fig.~1 (bottom left) in the main paper and the details of our proposed criterion.
	\item A probabilistic view of {\name} which decomposes the ambiguous distance between objects. 
	\item Detailed configurations of the experiments and implementation details.
	\item Additional experimental results to show the superiority of the proposed {\name}.
\end{itemize}
\end{abstract}

\section{Details of Our Proposed Criterion}
\subsection{Supervised CSL}
A supervised CSL model learns multiple embedding $\Psi_K$, revealing the specific characteristics of each similarity condition. 

Given $\Psi_K$, we can determine the validness of a triplet.
Specifically, given a (supervised) triplet $\tau=(\x,\y,\z, k)$ from the $k$-th condition, we use the corresponding embedding $\psi_k$ and compute the value 
\begin{align}
\mathbf{Diff}^k_\tau &= \mathbf{Dis}_{L_k}^2(\phi(\x), \;\phi(\z)) - \mathbf{Dis}_{L_k}^2(\phi(\x), \;\phi(\y))\notag\\
&=\|\psi_k(\x)-\psi_k(\z)\|_2^2 - \|\psi_k(\x)-\psi_k(\y)\|_2^2\;.\label{eq:diff}
\end{align}
We use $\mathbf{Diff}^k_\tau$ to predict whether a triplet is valid or not. if $\x$ (the anchor) and $\y$ (the target neighbor) are more similar than $\x$ (the anchor) and $\z$ (the impostor), \ie, the distance between $\x$ and $\y$ based on $\psi_k$ is smaller than the distance between $\x$ and $\z$, then we have $\mathbf{Diff}^k_\tau > 0$. 

\noindent{\bf Evaluation Criterion of Supervised CSL.}
To evaluate a supervised CSL model, we transform the task into measuring the prediction ability of multiple embeddings, \ie, we use the learned $\Psi_K$ to determine whether a given triplet is valid or not.
In detail, we sample the same number of {\em valid} triplets from each condition during evaluation. Then for each condition $k$, we use $\psi_k$ to predict whether triplets from that condition are correct or not based on the value of $\mathbf{Diff}^k_\tau$. 
The average accuracy (proportion of triplets predicted as correct) over them is used as the final criterion, which reveals the quality of multiple CSL embeddings. 

Denote $\mathbf{Map}[k]$ as a mapping from ground-truth condition $k$ to a particular learned embedding $\psi_{\mathbf{Map}[k]}$ in $\Psi_K$. In the supervised scenario, we learn the same number of conditional embeddings with the number of ground-truth conditions, and we set $\mathbf{Map}[k]=k$ as an identity mapping.
The evaluation steps are listed in Algorithm~\ref{alg:csl_evaluation}. 

It is notable that given a particular $\psi_k$, we have either $\mathbf{Diff}^k_\tau > 0$ or $\mathbf{Diff}^k_\tau \le 0$, so one triplet $\tau_1=(\x,\y,\z,k)$ and its reversed version $\tau_2=(\x,\z,\y,k)$ could not exist simultaneously under the same condition. Therefore, only using the {\em valid} triplets during evaluation (all of them with the ``correct'' label) produces reasonable results.

\begin{algorithm}[h]
	\caption{Evaluation of Supervised CSL}
	\begin{algorithmic}[1]	
		\Require Triplets $\{\tau_i\}_{i=1}^T$, the learned embeddings $\Psi_K$, the condition-embedding mapping $\mathbf{Map}[k]$\\
		\textbf{Initialize:} $t=0$.
		\For {$i=1 \to T$}
		\State 
		get $\tau_{i}=(\x,\y,\z,k)$ with ground-truth condition $k$
		\State compute $\mathbf{Diff}^k_{\tau_{i}}$ with $\psi_{\mathbf{Map}[k]}$ and Eq.~\ref{eq:diff}
		\If{$\mathbf{Diff}^k_{\tau_{i}}>0$}
		\State $t=t+1$
		\EndIf
		\EndFor\\
		\Return $Acc = t/T$
	\end{algorithmic}\label{alg:csl_evaluation}
\end{algorithm}

\noindent{\bf Discussions of the Supervised CSL Criterion.}
One direct question of the criterion is {\it whether it could reveal all conditional similarities or not}. Due to the fact that a triplet $\tau_1$ and its reversed version $\tau_2$ could not be satisfied by the same conditional embedding $\psi_k$, the evaluation of those valid triplets excludes the relationship revealed by the reversed triplets and characterize a particular similarity condition. The learned conditional embeddings could achieve high accuracy only when they characterize the latent relationship of all target conditions.

\subsection{Weakly Supervised CSL}
The current WS-CSL evaluation utilizes the same criterion as the supervised CSL~\cite{Tan2019Learning,Nigam2019Towards} --- predicting the correctness of a triplet {\em without using the test-time condition labels}. We first demonstrate the drawback of the current criterion via a synthetic experiment.

\noindent{\bf Synthetic Experiment Setups.}
We investigate various methods on UT-Zappos-50k dataset~\cite{Yu2014finegrained,Yu2017Semantic}, where we extract 40,000 triplets from four conditions. We use conditional labels for supervised methods, and neglect them for WS-CSL methods during evaluation. There are various methods in our synthetic experiment.
\begin{itemize}
    \item Optimal. We show the optimal {\it Supervised} CSL method as a reference. Which achieves the highest performance with the supervised evaluation criterion.
    \item CSN~\cite{Veit2017Conditional}. CSN learns multiple embedding masks during training in a fully {\em supervised} way, and applies the corresponding mask over the backbone as a special conditional measure. Note that the condition labels are required in both training and test processes. 
    \item LSN~\cite{Nigam2019Towards}. LSN is a multi-attribute model which discovers latent attributes by choosing the one with the minimum loss in a {\it Weakly Supervised} way. Following the multi-choice learning~\cite{Rivera2012Multiple}, it learns disentangled embeddings $\Psi_K$ with hard condition assignments.
    \item SCE-Net~\cite{Tan2019Learning}. SCE-Net is a {\it Weakly Supervised} method, which consists of an attention module and multiple embeddings $\Psi_K$. Given a triplet, SCE-Net generates a weight vector over multiple conditions, with which it fuses multiple embeddings $\Psi_K$ to predict the correctness of a triplet.
    \item {\name}. Our proposed {\it Weakly Supervised} CSL method with semantic regularization. Details can be found in the main paper.
\end{itemize}

We introduce another task to predict the correctness of reversed triplets, \ie, we transform $\tau_1=(\x,\y,\z,k)$ to $\tau_2=(\x,\z,\y,k)$ by exchanging the position of last two instances in the triplets. So in a supervised evaluation, the ground-truth of all reversed triplets must be false, and a supervised method should predict them as correct ones as fewer as possible (the lower the proportions of triplets be predicted as valid, the higher the accuracy).
However, in a WS-CSL evaluation, we do not use the condition label $k$, so a method predicts the correctness of $\tau_3=(\x,\y,\z)$ and $\tau_4=(\x,\z,\y)$ in the original and reversed cases directly. 

\begin{figure}[t]
	\centering
    \begin{minipage}[c]{0.95\linewidth}
	\includegraphics[width=\linewidth]{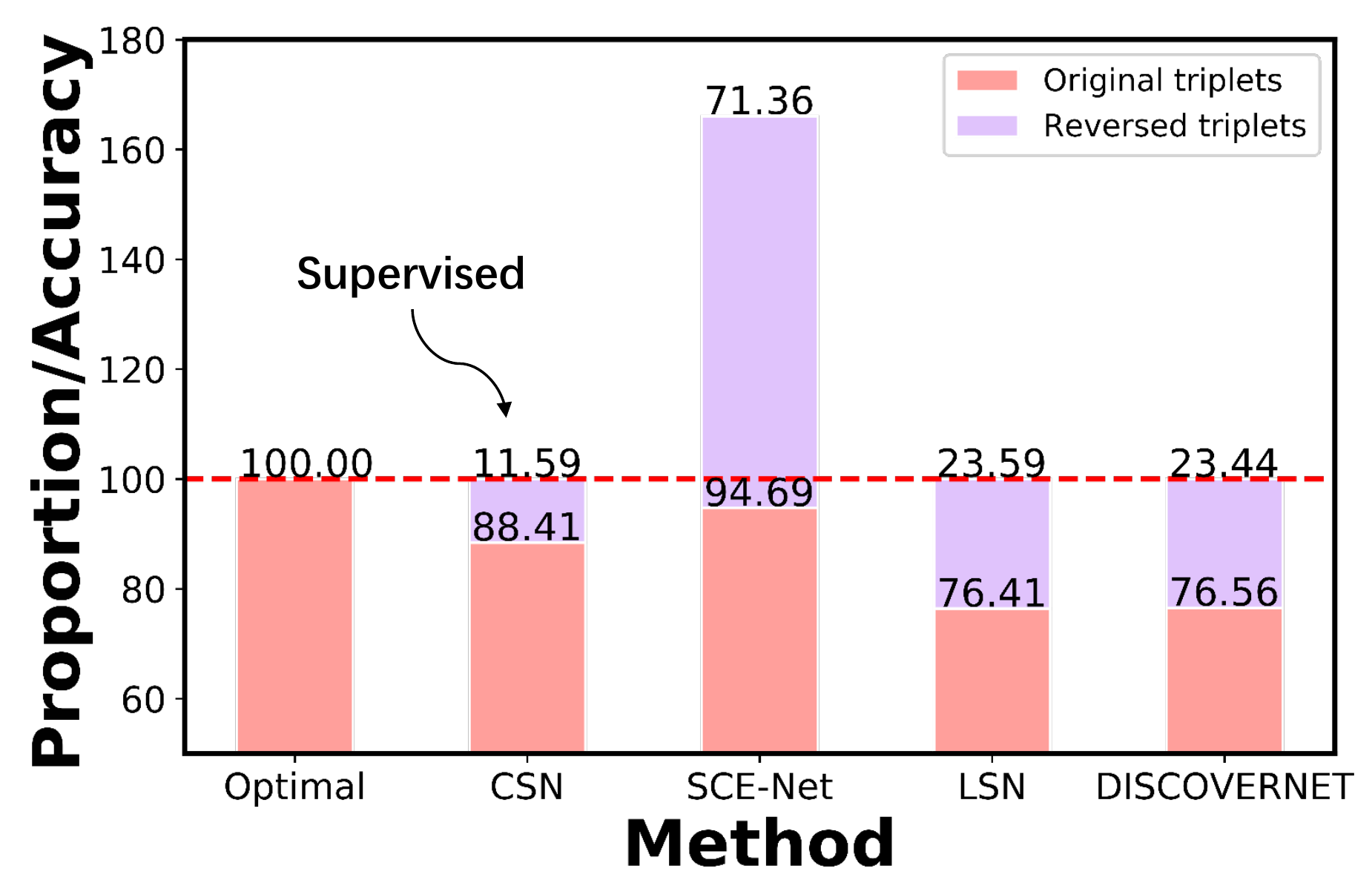}\\
    \end{minipage} 

	\caption{Given original (correct) triplets and their reversed variants on UT-Zappos-50k, we compute the proportion a model that predicts them as valid ones. The higher the proportion of {\em original} triplets be predicted as correct ones, the higher the accuracy. In contrast, the higher the proportion of {\em reversed} triplets be predicted as correct ones, the lower the accuracy.
	The first two are supervised CSL methods, and the last three are WS-CSL methods.	}\label{fig:teaser}
\end{figure}

\noindent{\bf Analysis of the Experiment.}
The results are shown in Fig.~\ref{fig:teaser}. We show the proportion of original and reversed triplets with red and blue respectively, and accumulate their values together. For the optimal supervised model at the leftmost, it predicts all original triplets as right ones and all reversed triplets as invalid (achieves 100\% accuracy in both cases). For the supervised CSN, it has high proportion (accuracy) on the original triplets and low proportion (also high accuracy) on the reversed ones. 

The phenomenon is different for WS-CSL methods, especially for SCE-Net. SCE-Net fuses all embeddings with attention for each triplet. The results show SCE-Net gets much higher accuracy on the original triplets (even better than the supervised CSN), but also predicts a lot of reversed triplets as correct ones. Fig.~\ref{fig:teaser} indicates that SCE-Net tends to treat most original and reversed triplets as right ones, which is different from the supervised case.

The main reason is that the original and reversed triplets could possess different conditions, so a WS-CSL method explains them from two diverse aspects. Using the supervised criterion (\ie, the red part) could be {\em biased} in this case. For example, SCE-Net is able to learn good embeddings and powerful attentions, but the current criterion demonstrates that it prefers valid triplets, which is explained by combined embeddings. Thus, we are unaware of whether we learn meaningful embeddings or a strong fusion module. 

\noindent{\bf Our Proposed WS-CSL Criterion.}
Based on our analysis, the supervised criterion is able to evaluate the quality of all learned embeddings, while the WS-CSL's criterion may fall into the scenario using all conditional embeddings to explain the triplets. We follow an intuitive way to measure a WS-CSL model --- whether the learned WS-CSL model performs similarly to the supervised CSL model. 
Then, not only the fusion of conditional embeddings $\Psi_K$ should cover all target semantics, but also the behavior of each $\psi_k$ reveals the relationship w.r.t. a specific condition.

We propose a new evaluation criterion (details could be found in the main paper). Using the conditional labels during evaluation, we find an alignment $\mathbf{Map}[k]$ between a particular condition and one of the learned embeddings in $\Psi_K$. The condition labels are only utilized to choose a good alignment. 
Then we can use the supervised criterion for a better evaluation. 
The details to evaluate with our criterion are listed in Algorithm~\ref{alg::ours}.

\begin{algorithm}[h]
	\caption{Our evaluation steps for WL-CSL methods}
	\label{alg::ours}
	\begin{algorithmic}[1]	
		\Require Learned embeddings $\Psi_K$, the number of ground-truth condition $K'$, the number of triplets of different ground-truth conditions $\{N_{k'}\}_{k'=1}^{K'}$.\\
		\textbf{\uppercase\expandafter{\romannumeral1}. Compute condition-embedding cost $C$.}
		\For {$k'=1 \to K'$}
		\For {$k=1 \to K $}
		\State 
		\textbf{Initialize:} $t=0$.
		\State get $\tau_{i}=(\x,\y,\z,k')$
		\State compute $\mathbf{Diff}^k_{\tau_{i}}$\;with $\psi_k$ and Eq.~\ref{eq:diff}
		\If{$\mathbf{Diff}^k_{\tau_{i}}>0$}
		\State $t=t+1$
		\EndIf
		\State $Acc_{k'k}=t/N_{k'}$
		\EndFor
		\EndFor\\
	    $C_{k'k} = 1 - Acc_{k'k}$\\
        \textbf{\uppercase\expandafter{\romannumeral2}. Match a condition $k'$ with embedding $\psi_k$.}\\
        \textbf{\romannumeral1. Greedy Alignment:}
		\For {$k'=1 \to K'$} 
		\State $\mathbf{Map}[k'] = \mathop{\arg\min}\limits_{k} C_{k'k}$
		\EndFor\\
		\textbf{\romannumeral2. OT Alignment:}\\
		compute transportation matrix T via\\{\small$
\min_{T\ge0} \langle T, C \rangle\qquad {\rm s.t.}\quad T{\mathbf 1} = \frac{1}{K}{\mathbf 1}, \;T^\top{\mathbf 1} = \frac{1}{K}{\mathbf 1}
$}
		\For {$k'=1 \to K'$} 
		\State $\mathbf{Map}[k'] = \mathop{\arg\max}\limits_{k} T_{k'k}$
		\EndFor\\
	    \textbf{\uppercase\expandafter{\romannumeral3}. Compute accuracy with Alg.~\ref{alg:csl_evaluation} and $\mathbf{Map}[k]$.}
	\end{algorithmic}
\end{algorithm}

\noindent\textbf{Discussions.}
A natural question for obtaining the conditional alignment is why we use OT instead of the Hungarian method.
OT is a more {\em general} method to solve a matching problem. 
OT can deal with the case when there exists a number mismatch between the conditional embeddings and ground-truth conditions. 
We can show that when we use uniform marginal distributions and a square cost matrix in OT, the transportation will degenerate to the same solution as Hungarian's output with special conditions~\cite{PeyreC19OT}.

\section{Probabilistic View of {\scshape DiscoverNet}}
In Weakly Supervised Conditional Similarity Learning (WS-CSL), only triplets $\{\tau=(\x,\;\y,\;\z)\}$ from multiple conditions are provided, and the condition label $k$ in each triplet is {\em unknown}. This is the usual case that we can explicitly describe the similarity of two images but difficult to point out from which aspect we measure them clearly.
The model needs to infer the right condition label and learn discriminative embeddings to capture objects' characteristics simultaneously.
Here we provide a detailed description of the probabilistic view of the basic version of {\name}. The final {\name} is equipped with a set module or a semantic regularizer.

\begin{table*}[tbp]
	\centering
	\caption{The detailed configurations of conditions in our synthesized Celeb-A$^\dagger$. We synthesize 5 attributes over Celeb-A via combining similar binary attributes together, so each condition has 5-7 possible values.} 
	\begin{tabular}{c|c|c}
		\addlinespace
		\toprule
		Combined attributes & \# of values & Original attributes included \\
		\midrule
		Hair color & 5 & black-hair,blond-hair,brown-hair,gray-hair\\
		\midrule
		Hair type & 7 & bangs,receding-hairline,straight-hair,wavy-hair\\
		\midrule
		Eye and eyebone & 6 & arched-eyebrows,bags-under-eyes,bushy-eyebrows,narrow-eyes\\
		\midrule
		Accessories & 6 & wearing-earrings,wearing-hat,wearing-necklace,wearing-necktie\\
		\midrule
		Nose and mouth & 7 & big-lips,big-nose,mouth-slightly-open,pointy-nose\\
		\bottomrule
	\end{tabular}\label{table:combineattrs}
\end{table*}

{\name} characterizes both the {\em instance-instance} and {\em triplets-condition} relations in a ``decompose-and-fuse'' manner, which could be interpreted in a probabilistic aspect. With a bit abuse of the notation, we use $\tau=1$ to represent the comparison relationship in the triplet is true, and $\tau=0$ otherwise. With the help of the latent variable $c_\tau$, we can measure the validness of a triplet $\tau$ via a holistic consideration of all similarity conditions:
\begin{equation}
\Pr(\tau=1) =\; \prod_{k=1}^K \Pr(\tau=1|c^k_\tau=1)^{c^k_\tau}
\;.\label{eq:conditional}
\end{equation}
$c_\tau\in\{0,1\}^K$ is a multinomial random variable, and we use  $c^k_\tau=1$ to denote the $k$-th component is selected. $\sigma(x) = \frac{1}{1+\exp(-x)}$ is the sigmoid function, which squashes a variable into the range [0,1]. 
In Eq.~\ref{eq:conditional}, whether the relationship in the triplet is true or not depends both on the activated latent condition $c_\tau^k$ and the probability of the triplet in that view $\Pr(\tau=1|c^k_\tau=1)$.
Thus, the probability is related to both the concept prior $c^k_\tau=1$ (the ``triplets-condition'' relationship) and the validness of the triplet conditioned on a particular concept $\Pr(\tau=1|c^k_\tau=1)$ (the ``instance-instance'' relationship). 
The influence of all similarity conditions is aggregated together to determine the triplet in expectation.

Based on the projection $L_k$, we can define the probability that a given triplet is true based on their distance difference in the specific instance-instance embedding space
\begin{equation}
\Pr(\tau=1 | c_\tau^k=1) = \sigma\left(\mathbf{Diff}^k_\tau - \gamma\right).
\end{equation}
where the distance difference $\mathbf{Diff}^k_\tau$ is defined in Eq.~\ref{eq:diff}.
If the distance between $\x$ and $\y$ in this embedding space based on $\phi$ is smaller than the distance between $\x$ and $\z$, the input to $\sigma$ is large such that the triplet will have a large probability to be valid. 
$\gamma > 0$ is a threshold. By subtracting $\gamma$ from $\mathbf{Diff}^k_\tau$, we require the distance with the impostor should not only be large but also larger than the distance with the target neighbor plus a margin.

For all given triplets $\mathcal{T}$, we optimize the embedding by maximizing the log-likelihood:
\begin{equation}
\mathcal{O} = \sum_{\tau\in\mathcal{T}} \log \Pr(\tau=1)\;,\label{eq:prob_obj}
\end{equation} 
which has a similar form of the large margin loss~\cite{Schroff2015FaceNet,Song2016Deep} when only the general projection $L$ and $\phi$ is used. The overall objective which minimizes the negative log likelihood of the joint probability over all triplets is:
\begin{align}
\mathcal{O} &=\; -\sum_{\tau\in\mathcal{T}} \log \prod_{k=1}^K\sigma(\mathbf{Diff}^k_\tau - \gamma)^{c_\tau^k}\label{eq:objective}\\
&=\; -\sum_{\tau\in\mathcal{T}}\sum_{k=1}^K c_\tau^k \log \sigma(\mathbf{Diff}^k_\tau - \gamma)\notag\\
&=\; -\sum_{\tau\in\mathcal{T}}\mathbb{E}_{c_\tau} \log \sigma(\mathbf{Diff}^k_\tau - \gamma)\notag\\ &=\;\sum_{\tau\in\mathcal{T}}\mathbb{E}_{c_\tau} \ell(\mathbf{Diff}^k_\tau - \gamma)\notag\;\\
&\approx\;\sum_{\tau\in\mathcal{T}}\ell\left(\mathbb{E}_{c_\tau}\left[\mathbf{Diff}^k_\tau\right] - \gamma\right)\notag\;.
\end{align}
Here $\ell(x) = \log (1+\exp(-x))$ is the logistic loss function, which can be replaced by other general losses.
The approximation transforms the expectation over the loss function to the {\em expected distance over the conditions}, which fuses and reveals the preference over multiple similarity conditions. 
Therefore, the optimization in Eq.~\ref{eq:objective} requires the relationship in the selected instance-wise metric space to reveal the corresponding similarity condition, which makes the distance between the anchor and the impostor larger than the distance between the anchor and the target neighbor.

\section{Experimental Setups}
We describe the dataset and the implementation details in this section.
\subsection{Datasets}
\textbf{Celeb-A$^\dagger$} is a more complicated version of Celeb-A~\cite{Liu2015Deep} via combining similar binary attributes in Celeb-A together. In particular, there are 202,599 face images from different identities in Celeb-A, and 40 binary visual attributes for each image, \eg, ``Eyeglasses'' or ``Wearing Hat''. Each attribute corresponds to a condition, and we sample triplets randomly based on the binary values for each condition. To increase the difficulty of the dataset, we combine related binary attributes in Celeb-A together. The attributes related to ``Hair color'', ``Hair type'', ``Eye and eye-bone'', ``Accessories'', ``Nose and mouth'' are merged, and each of them has 5-7 possible discrete values. Details can be found in Table~\ref{table:combineattrs}. Thus, different from the vanilla version with 40 binary attributes, the smaller number of attributes in the synthesized new dataset \textbf{Celeb-A$^\dagger$} has multi-choice conditions. We apply the same model configurations (such as the learning rate and the architecture) for Celeb-A$^\dagger$ as Celeb-A.

\subsection{Implementation Details}
Following~\cite{Veit2017Conditional,Nigam2019Towards,Tan2019Learning}, we use ResNet-18~\cite{He2016Residual} to implement the embedding backbone $\phi$. Different from the previous literature fine-tuning the backbone based on the weights pre-trained on ImageNet~\cite{Deng2009Imagenet}, we also consider the case that we train the full model from scratch. We find although the pre-trained weights make the model predict triplet well, it losses the coverage of semantics among conditions.
The last downsampling layer in the backbone is removed to accommodate the smaller image size, and an additional fully connected layer is appended to project the embeddings to specified dimensions. 
We set the embedding dimension 64 and the temperature $\varsigma$ in Eq. 10 in the main paper as 1.0 for both UT-Zappos-50k and Celeb-A. 
We use the Adam optimizer~\cite{Kingma2015Adam} in our experiments and train our model for 90 epochs totally. The initial learning rate is 0.01 and annealed to 10\% every 30 epochs. When the model is fine-tuned from the pre-trained weights, we set the initial learning rate as 5e-4, and the learning rate of the last layer is 10 times faster. In the case of training a model from scratch, the initial learning rate is 0.01. Other configurations are the same as the scenario optimized over the pre-trained weights.  
Margin loss is used to optimize the embedding, and each mini-batch contains 64 triplets. 

\section{Additional Experiments}
In this section, we further investigate our proposed {\name}, including additional benchmark evaluations, some ablation studies, and several visualizations omitted in the main paper.

\subsection{Additional Benchmark Evaluations}
\begin{table}[tbp]
	\centering
	\caption{Greedy accuracy and OT accuracy on 8-condition Celeb-A (binary conditions) and its attribute merged variant Celeb-A$^\dagger$ with five multi-choice conditions, respectively. All methods are fine-tuned with pre-trained weights.
	We make the best WS-CSL results in bold.} 
	\tabcolsep 2pt
	\begin{tabular}{c|cc||cc}
		\addlinespace
		\toprule
		& \multicolumn{2}{c||}{Celeb-A} & \multicolumn{2}{c}{Celeb-A$^\dagger$}\\
		\midrule
		Criteria $\rightarrow$ & GR Acc. & OT Acc. & GR Acc. & OT Acc. \\
		\midrule
		CSN~\cite{Veit2017Conditional}  & 84.88 & 84.88 & 73.04 & 73.04 \\
		LSN~\cite{Nigam2019Towards} & 72.89 & 71.95 & 63.73 & 63.67\\
		SCE-Net~\cite{Tan2019Learning}	 & 69.91 & 68.73 & 60.26 & 59.73\\
		\midrule
		{\name}$_{\rm Set}$ & {\bf 80.65} & {\bf 78.81} & 64.23 & 63.49\\ 
		{\name}$_{\rm Reg}$ & 79.31 & 78.79 & {\bf 65.32} & {\bf 64.71}\\
		\bottomrule
	\end{tabular}\label{table:celebra}
\end{table}

\noindent{\bf Celeb-A.} The results of all methods over Celeb-A variants are listed in Table~\ref{table:celebra}. 
We observe that {\name} variants get the best accuracy with greedy or OT alignments among other WS-CSL methods.

\begin{table}[tbp]
	\centering
	\caption{Influence of the number of training triplets for {\name}$_{\rm Set}$ and {\name}$_{\rm Reg}$ on Celeb-A. Models are fine-tuned with pre-trained weights.} 
	\tabcolsep 3pt
	\begin{tabular}{c|cc||cc}
		\addlinespace
		\toprule
		& \multicolumn{2}{c||}{{\name}$_{\rm Set}$} & \multicolumn{2}{c}{{\name}$_{\rm Reg}$}\\
		\midrule
		\# Number & GR Acc. & OT Acc. & GR Acc. & OT Acc. \\
		\midrule
		$1\times10^5$  & 79.08 & 78.07 & 78.18 & 75.66 \\
		$2\times10^5$  & 79.83 & 77.55 & 78.28 & 76.14\\
		$4\times10^5$  & 80.65 & 78.81 & 79.31 & 78.79\\
		\bottomrule
	\end{tabular}\label{tab:ntriplets}
\end{table}

\subsection{Ablation Studies}
\begin{table}[tbp]
	\centering
	\caption{Performance comparison between {\name}$_{\rm Tra}$ (the Transformer~\cite{Vaswani2017Attention} implementation of $g$) and {\name}$_{\rm Set}$ on UT-Zappos-50k. We investigate two cases that training the model from scratch and fine-tune the model with pre-trained weights. Both GR accuracy and OT accuracy are measured.}
	\tabcolsep 2pt
	\begin{tabular}{c|cc||cc}
		\addlinespace
		\toprule
		Setups $\rightarrow$ & \multicolumn{2}{c||}{w/ pretrain} & \multicolumn{2}{c}{w/o pretrain}\\
		\midrule
		Criteria $\rightarrow$ & GR Acc. & OT Acc.& GR Acc. & OT Acc. \\
		\toprule
		{\name}$_{\rm Set}$ & 76.98 & 75.68 &  74.67 &  74.13\\
		{\name}$_{\rm Reg}$ & {\bf 77.84} & {\bf 77.68} & 72.99 & 71.46\\
		\midrule
		{\name}$_{\rm Tra}$ & 77.54 & 77.30 & {\bf 75.72} & {\bf 75.46}\\
		\bottomrule
	\end{tabular}\label{table:trans}
\end{table}

\noindent\textbf{Another choice of the set module}.
In {\name}$_{\rm Set}$, we use the maximum operator to make the output of the pair sets in a triplet become permutation invariant~\cite{Zaheer2017Deepsets}. 
Inspired by~\cite{Ye2020Few}, we can also investigate the mapping function $g$ with Transformer~\cite{Vaswani2017Attention}.
The multi-head self-attention mechanism keeps the set property of the mapping but improves the learning ability. 
After replacing $g$ with the Transformer, we name the variants of set module as {\name}$_{\rm Tra}$. 
We compare the performance of {\name} variants on the Zappos dataset. 
As shown in Table~\ref{table:trans}, {\name}$_{\rm Tra}$ outperforms {\name}$_{\rm Set}$ by a large margin if the model is optimized without the pre-trained weights.

\noindent\textbf{Influence of training triplets.}
We show the influence of the triplets number in Table~\ref{tab:ntriplets}, where we vary the number of triplets during the training progress of {\name}. 
More training triplets make it easier for a model to infer the latent conditions of triplets and shape the various spaces.
As shown in Table~\ref{tab:ntriplets}, when there are more training triplets, both the GR accuracy and OT accuracy increase. The results also indicate that {\name} is able to learn discriminative conditional embeddings and identify the latent conditions given a relatively small number of training triplets. 

\noindent\textbf{Influence of the balance weight $\lambda$ of semantic regularization.} Table~\ref{tab:ratio} shows the influence of $\lambda$ in {\name}$_{\rm Reg}$. We find that {\name}$_{\rm Reg}$ can get higher accuracy when $\lambda$ increases around 0.001, which indicates the regularization indeed helps.

\begin{table}[tbp]
	\centering
	\caption{Influence of the balance weight $\lambda$ over the training regularizer for {\name}$_{\rm Reg}$ on Celeb-A. Models are fine-tuned with pre-trained weights.} 
	\begin{tabular}{c|c|c|c|c|c}
		\addlinespace
		\toprule
		$\lambda$ & 0 & 0.0001 & 0.001 & 0.01 & 0.1\\
		\midrule
		GR Acc. & 73.81 & 77.51 & 77.70 & 75.19 & 50.87\\
		OT Acc. & 69.52 & 76.19 & 77.09 & 74.07 & 50.46\\
		\bottomrule
	\end{tabular}\label{tab:ratio}
\end{table}

\begin{figure*}
	\centering
	\begin{minipage}[h]{0.48\linewidth}
		\centering 
		\includegraphics[width=\textwidth]{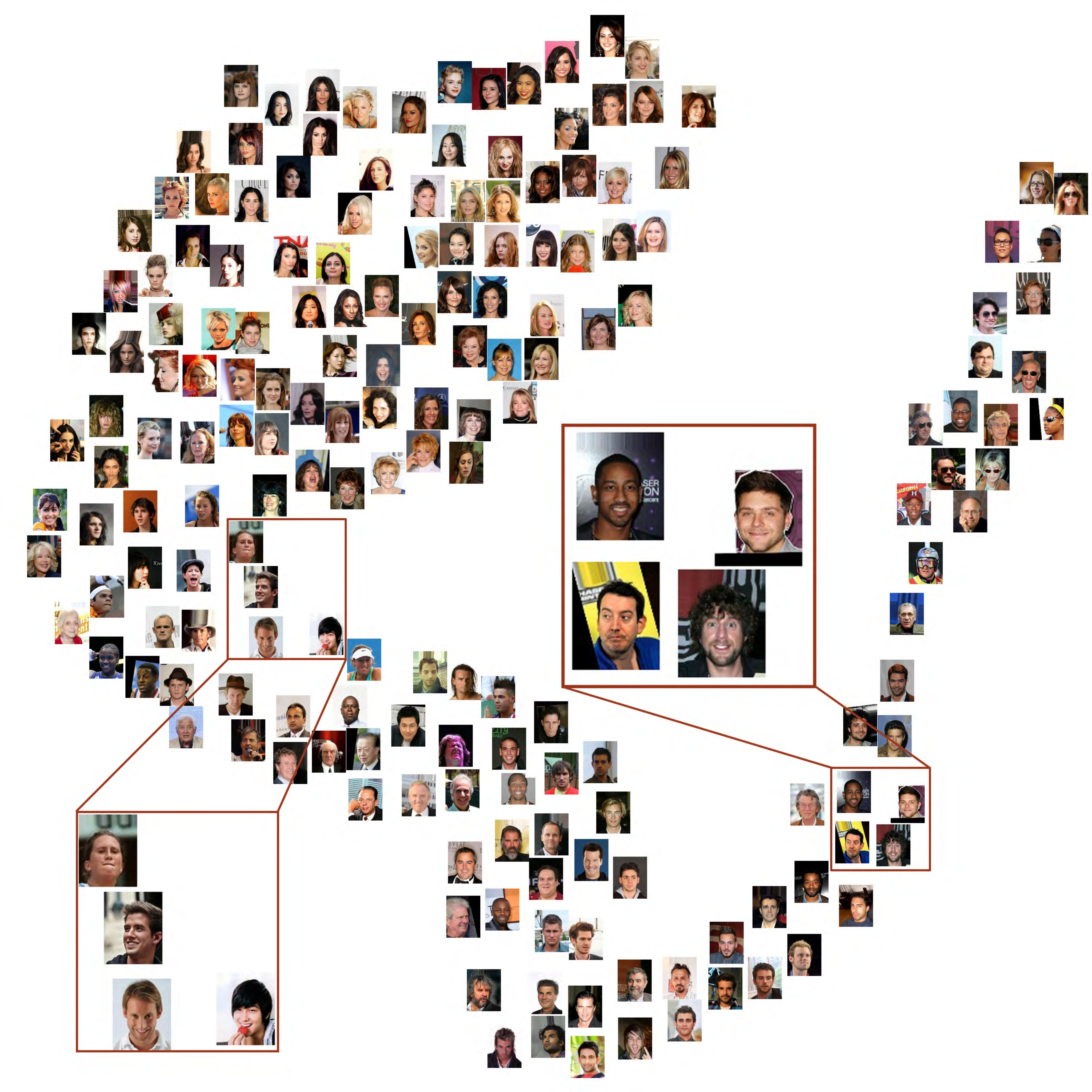}\\
		\mbox{({\it a}) {5-o-Clock-Shadow}}
	\end{minipage}
	\begin{minipage}[h]{0.48\linewidth}
		\centering 
		\includegraphics[width=\textwidth]{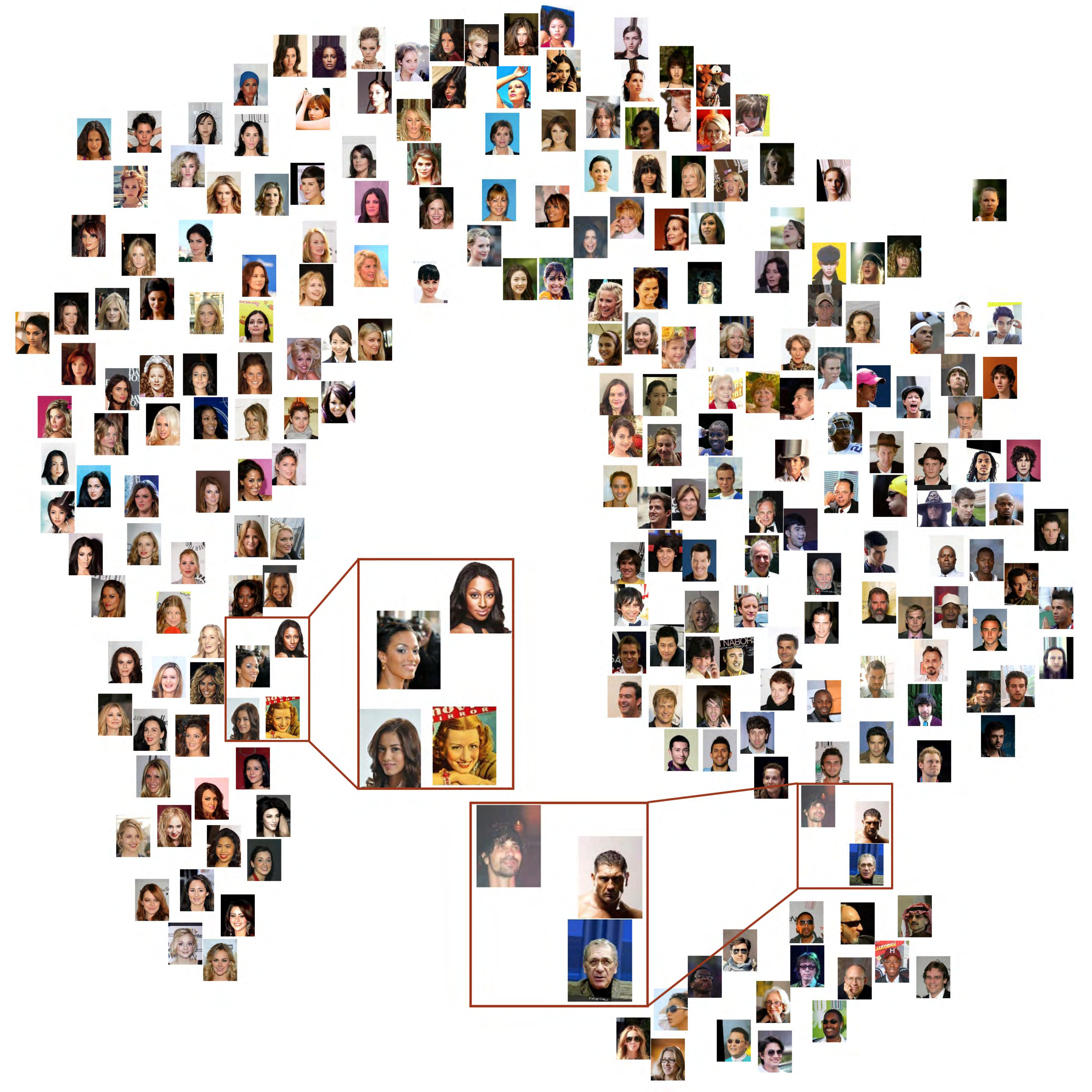}\\
		\mbox{({\it b}) {Attractive}}
	\end{minipage}
	\begin{minipage}[h]{0.48\linewidth}
		\centering 
		\includegraphics[width=\textwidth]{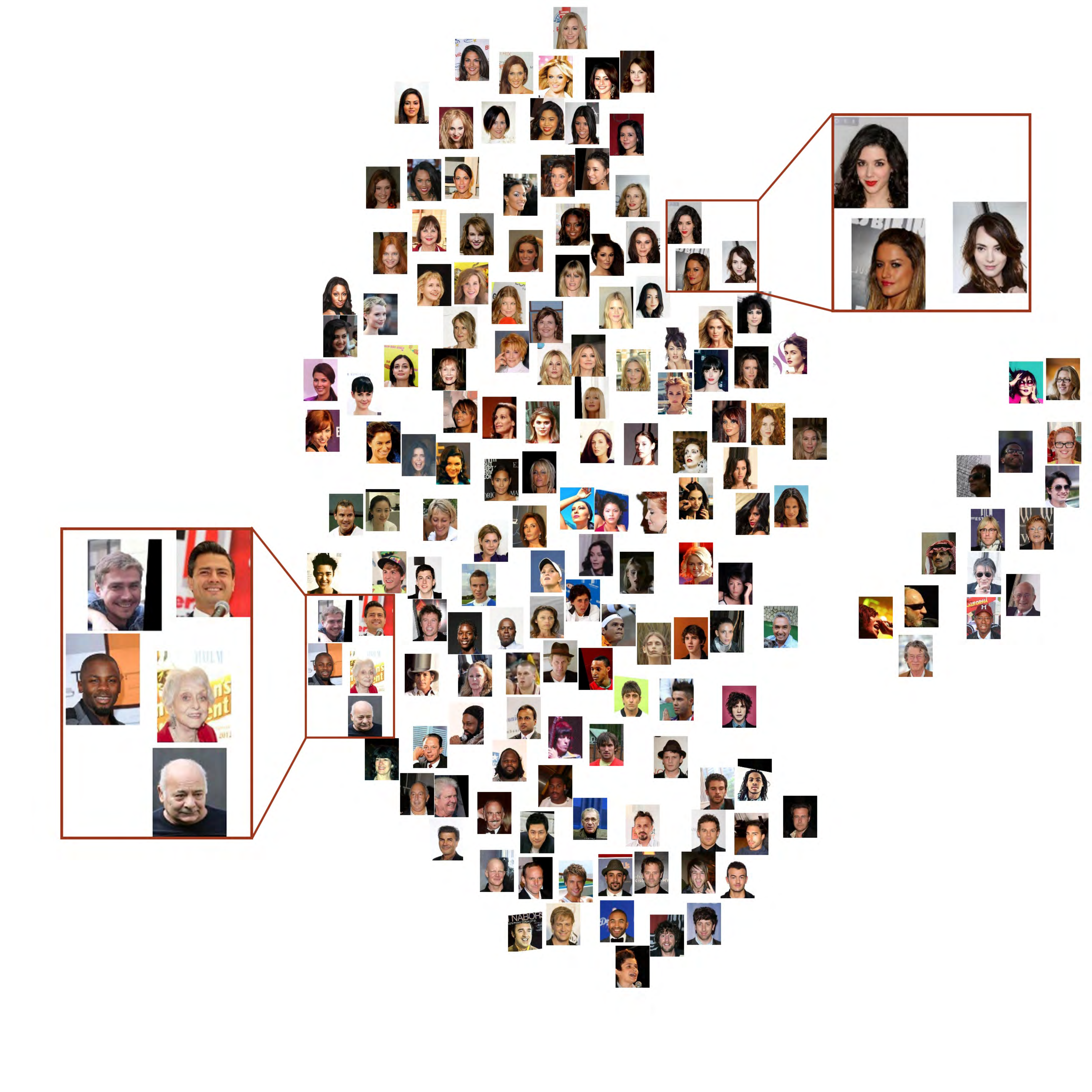}\\
		\mbox{({\it c}) {Bags-Under-Eyes}}
	\end{minipage}
	\begin{minipage}[h]{0.48\linewidth}
		\centering 
		\includegraphics[width=\textwidth]{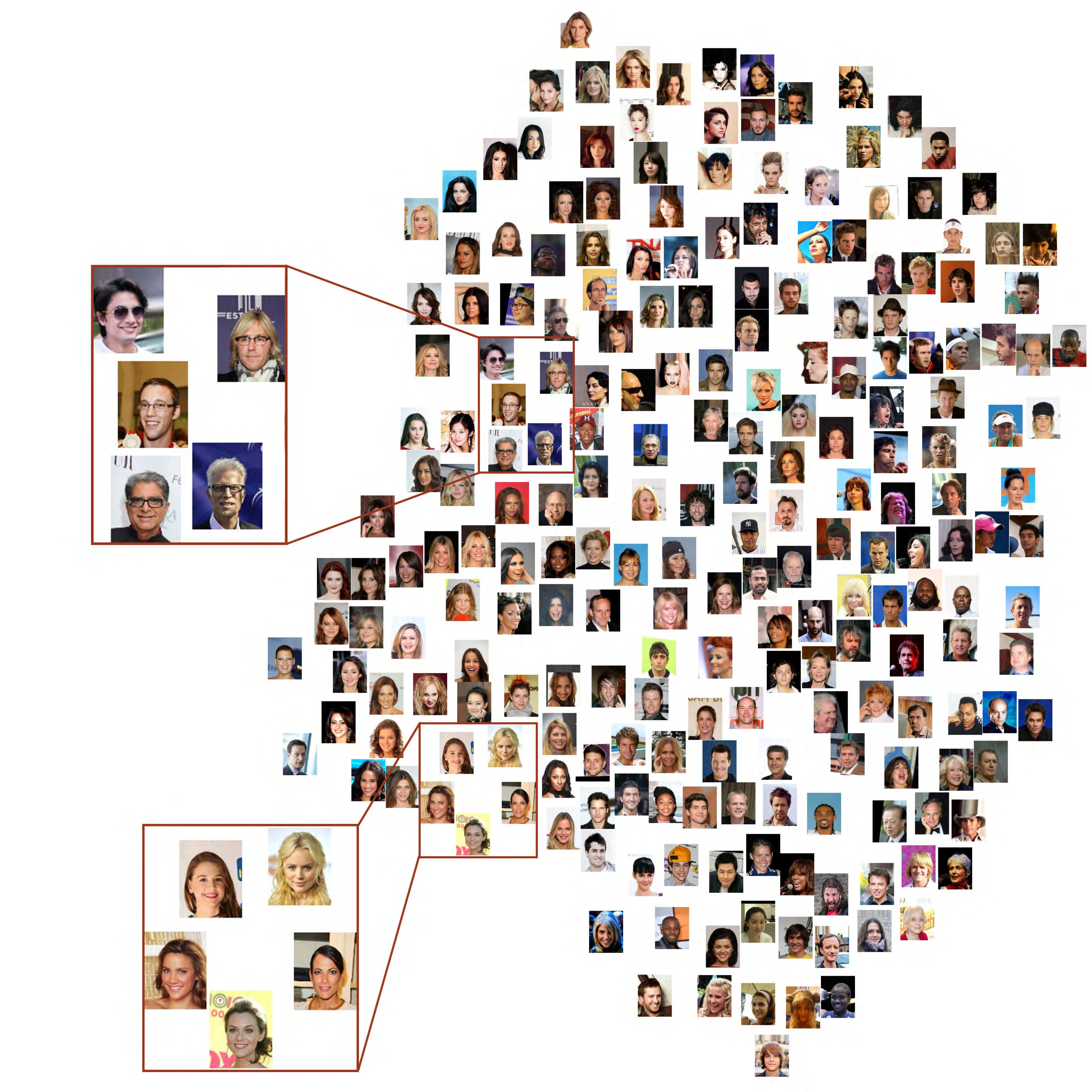}\\
		\mbox{({\it d}) {Eyeglasses}}\label{fig:TSNE-1-d}
	\end{minipage}
	\caption{TSNE of the learned embeddings for each of the four conditions (\ie, 5-o-Clock-Shadow, attractive, bags-under-eyes and eyeglasses) on Celeb-A dataset based on {\name}$_{\rm Reg}$.}\label{fig:TSNE-1}
\end{figure*}

\begin{figure*}
	\centering
	\begin{minipage}[h]{0.48\linewidth}
		\centering 
		\includegraphics[width=\textwidth]{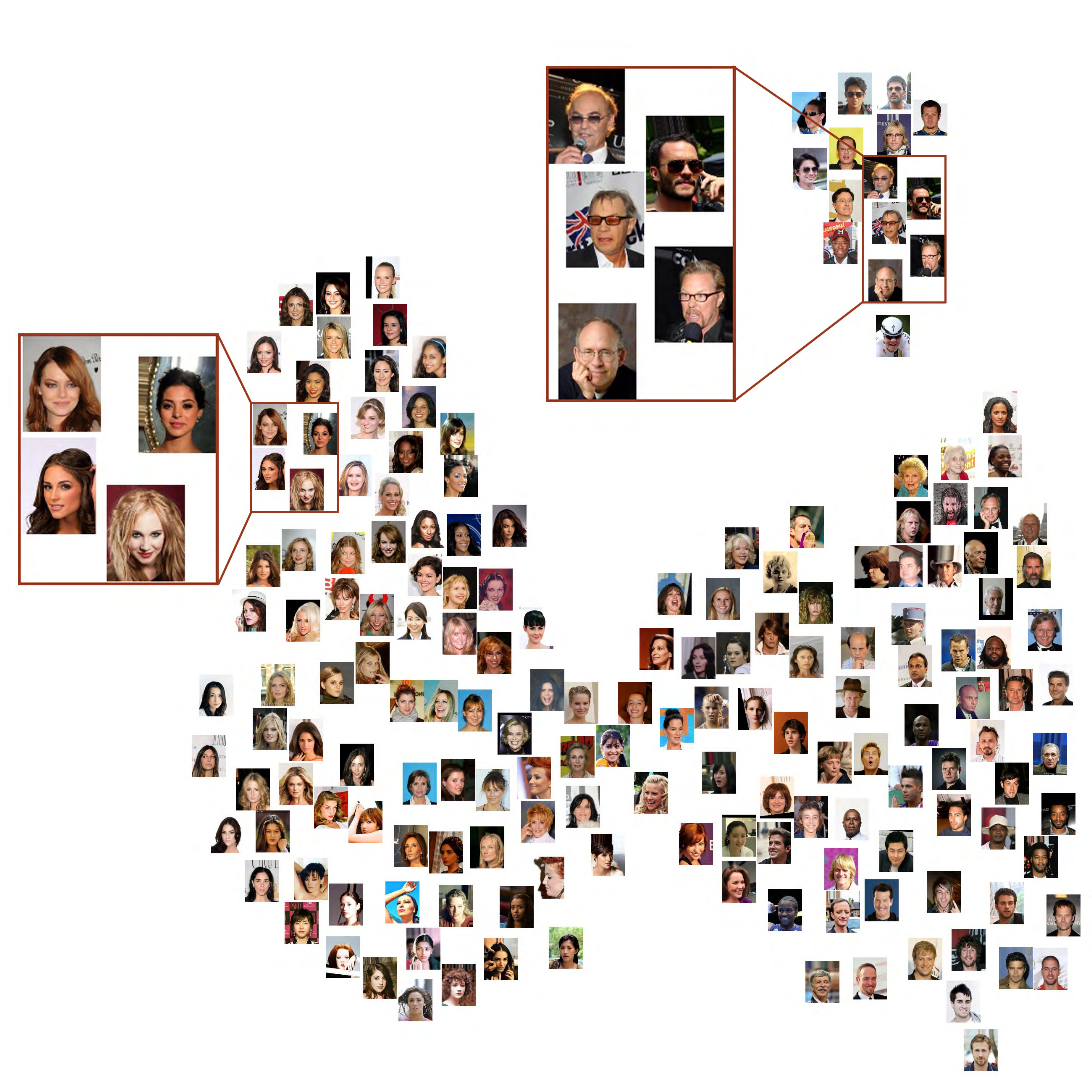}\\
		\mbox{({\it a}) {Male}}
	\end{minipage}
	\begin{minipage}[h]{0.48\linewidth}
		\centering 
		\includegraphics[width=\textwidth]{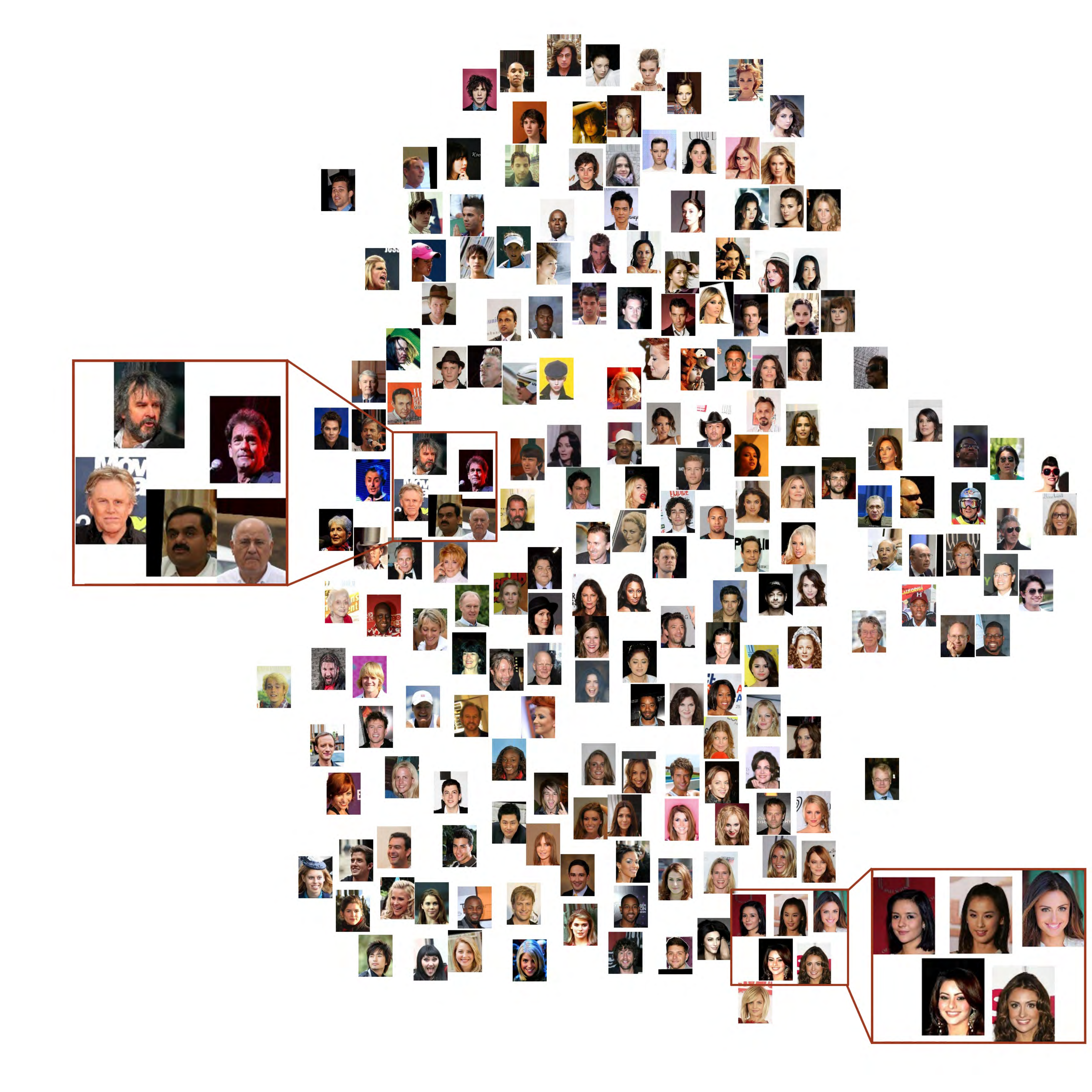}\\
		\mbox{({\it b}) {Smiling}}
	\end{minipage}
	\begin{minipage}[h]{0.48\linewidth}
		\centering 
		\includegraphics[width=\textwidth]{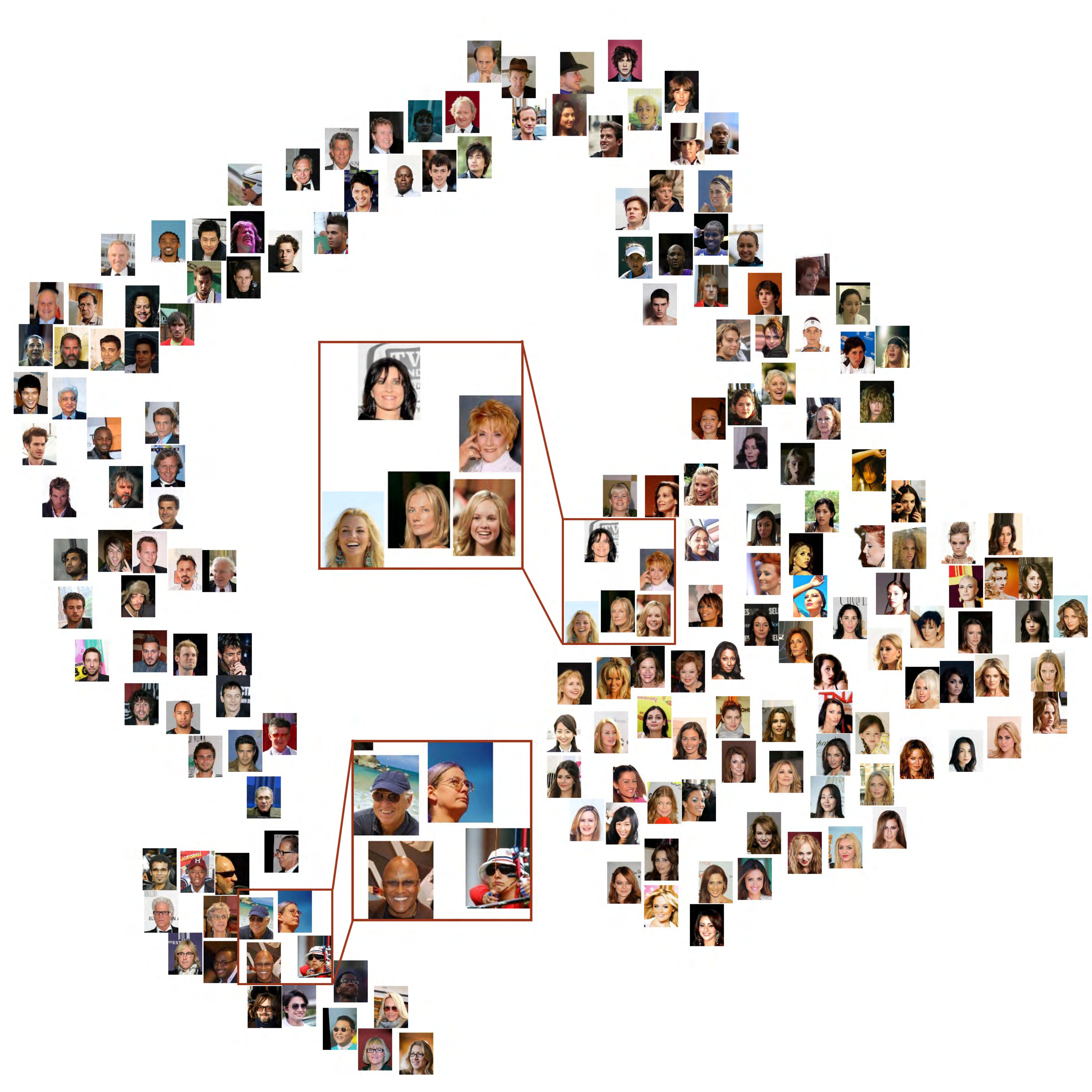}\\
		\mbox{({\it c}) {Wearing-Lipstick}}
	\end{minipage}
	\begin{minipage}[h]{0.48\linewidth}
		\centering 
		\includegraphics[width=\textwidth]{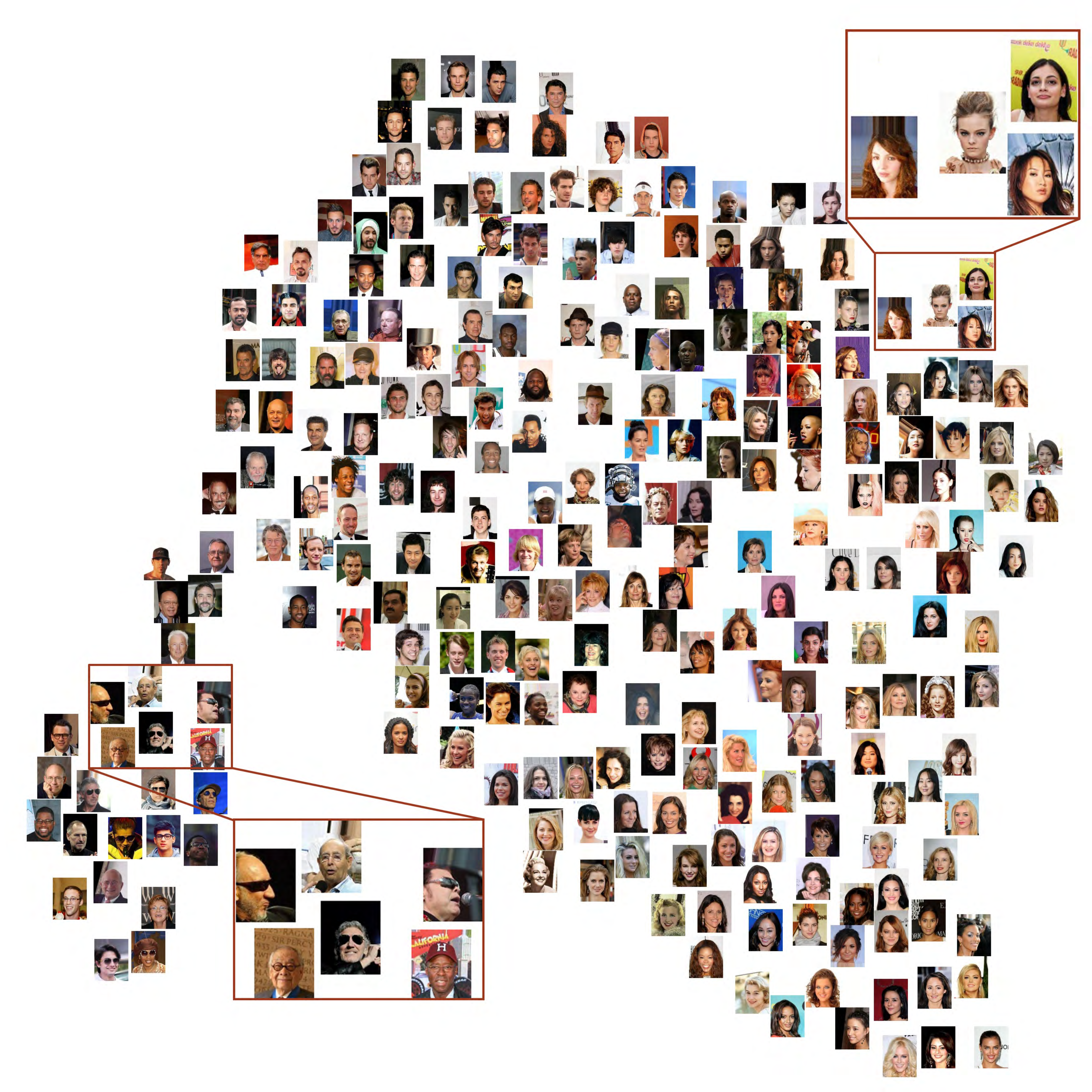}\\
		\mbox{({\it d}) {Young}}
	\end{minipage}
	\caption{TSNE of the learned embeddings for each of the four conditions (\ie, male, smiling, wearing-lipstick, and young) on Celeb-A dataset based on {\name}$_{\rm Reg}$.}\label{fig:TSNE-2}
\end{figure*}

\paragraph{Visualization of semantic embeddings.}
To better illustrate the ability that {\name} learns meaningful semantics given the triplets, we show the TSNE visualizations for each of the learned semantic spaces by {\name}$_{\rm Reg}$ on Celeb-A in Fig.~\ref{fig:TSNE-1} and Fig.~\ref{fig:TSNE-2}. Obviously, {\name} captures the variety of conditions and learns different embeddings for the dataset with good interpretability. Typically, in Fig.~\ref{fig:TSNE-1} (d), on the ``eyeglasses'' condition, {\name} gathers all faces with eyeglasses in the center-left part of the image. Similarly, in Fig.~\ref{fig:TSNE-2} (b), on the ``smiling'' condition, {\name} gathers all faces without smile in the upper-left part of the image. Note that {\name} learns the semantic metric spaces without any condition labels.

\begin{figure*}
	\centering
	\begin{minipage}[h]{\linewidth}
		\centering 
		\includegraphics[width=\textwidth]{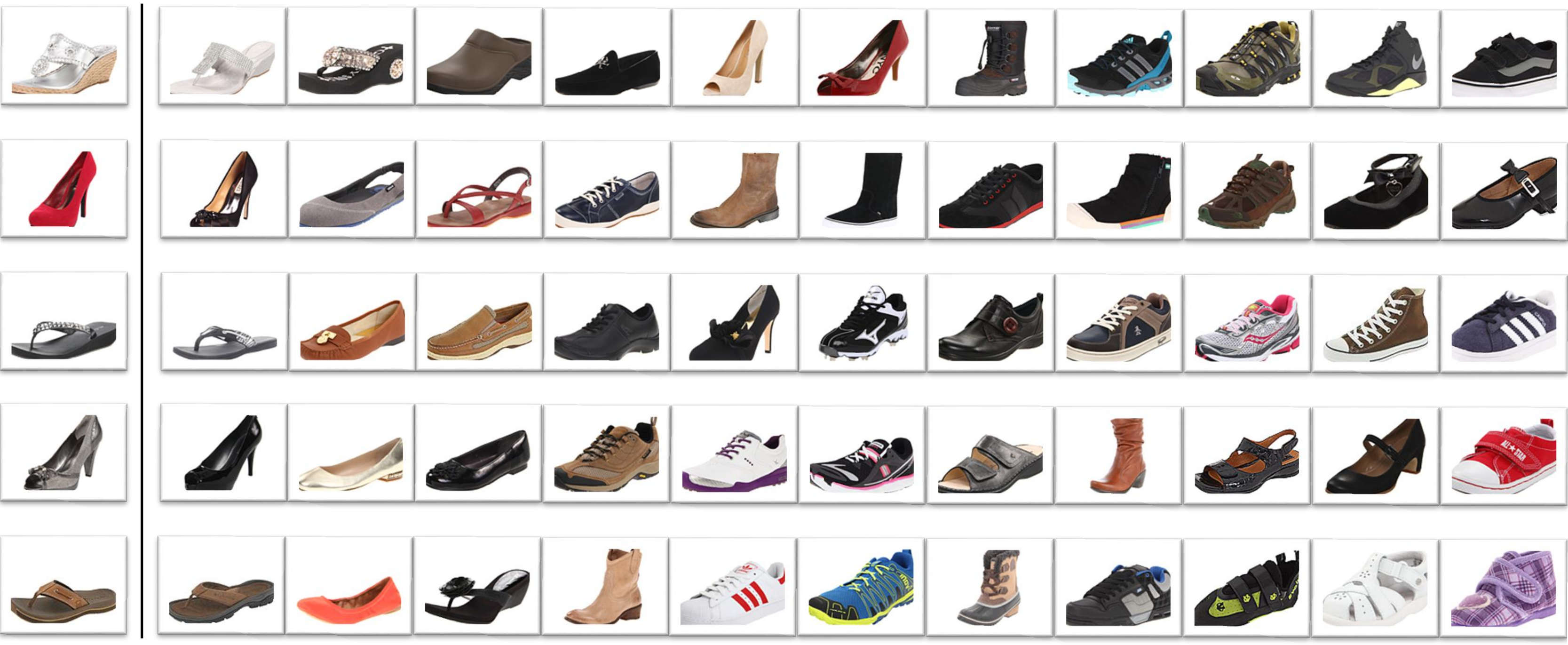}\\
		\mbox{({\it a}) {Functional Types}}
	\end{minipage}
	\begin{minipage}[h]{\linewidth}
		\centering 
		\includegraphics[width=\textwidth]{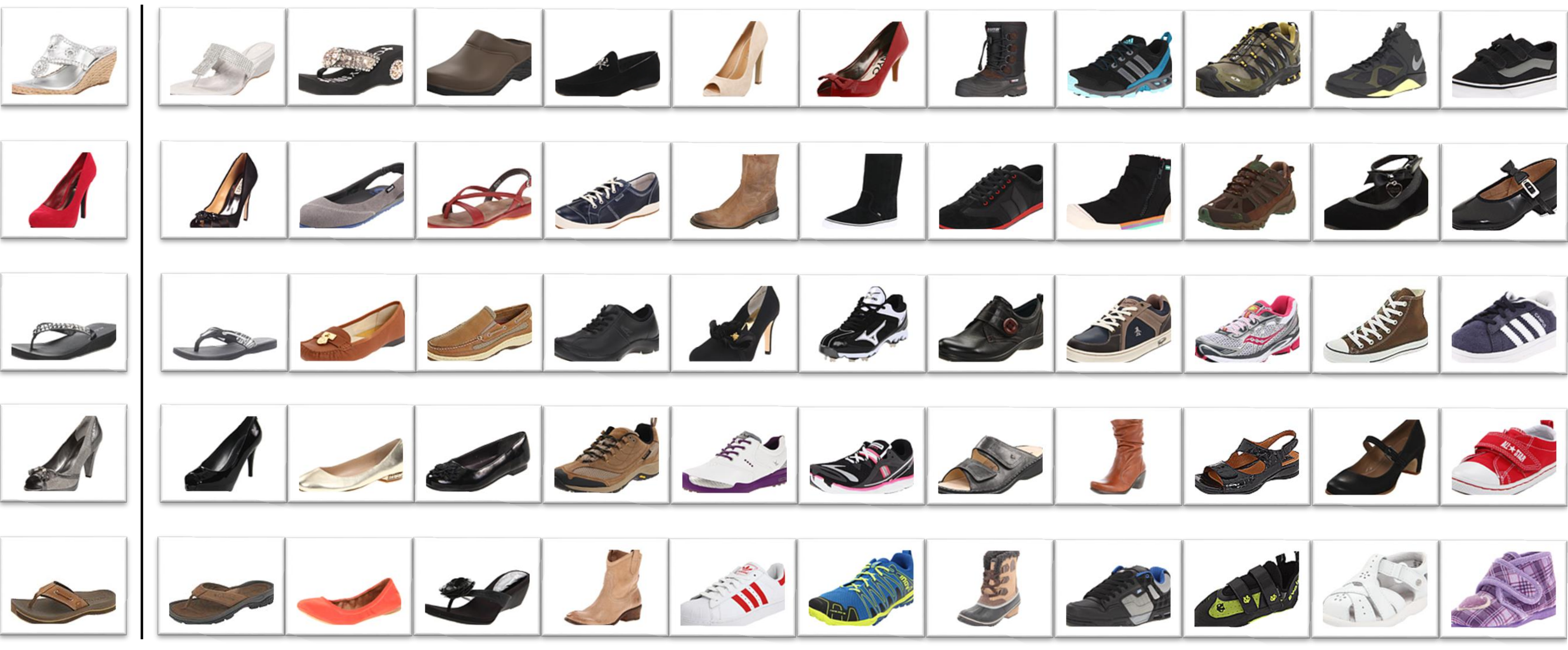}\\
		\mbox{({\it b}) {Closing Mechanism}}
	\end{minipage}
	\caption{Visualization of the image retrieval results for each of the two conditions (\ie, functional types and closing mechanism) on UT-Zappos-50k dataset with the learned embedding of {\name}$_{\rm Reg}$. The first image in each row is the query item, and shoes are ranked by distances to the query item in ascending order.}\label{fig:retrieval-1}
\end{figure*}

\begin{figure*}
	\centering
	\begin{minipage}[h]{\linewidth}
		\centering 
		\includegraphics[width=\textwidth]{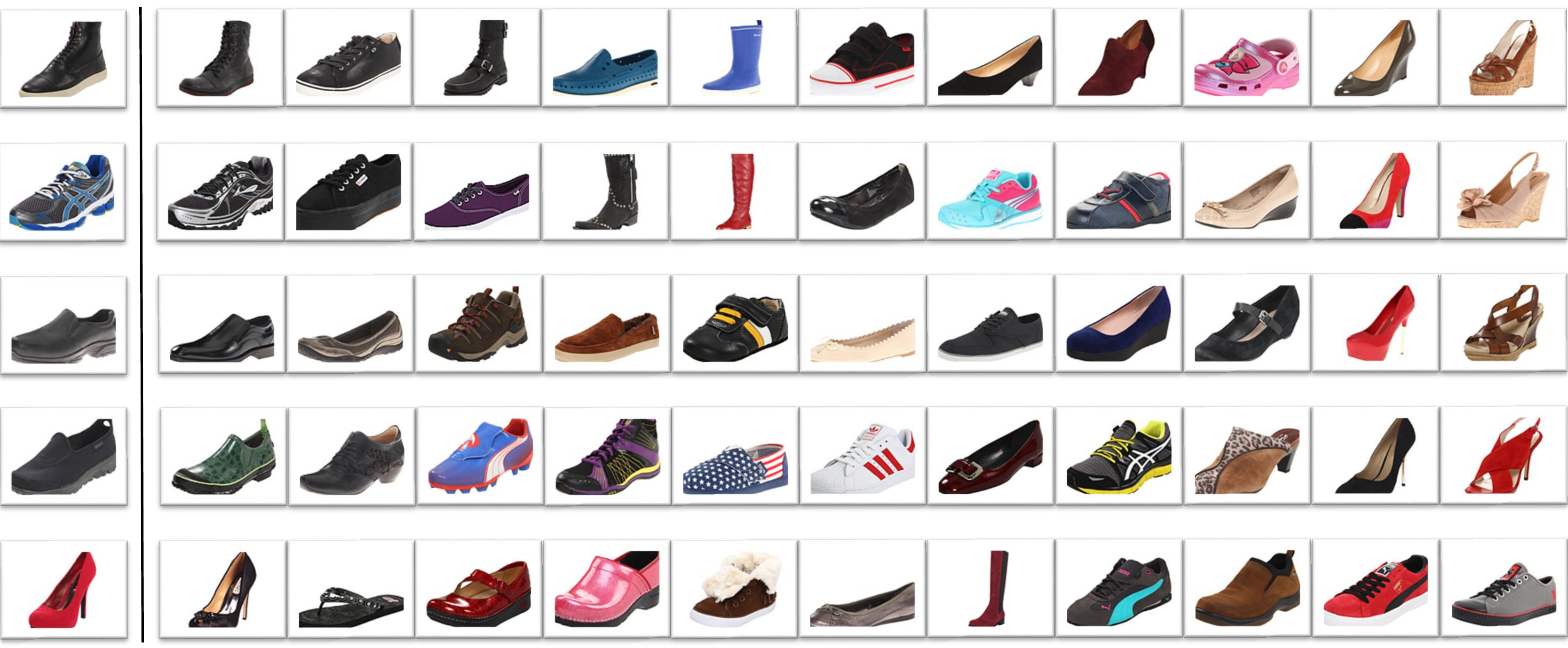}\\
		\mbox{({\it a}) {Suggested Gender}}
	\end{minipage}
	\begin{minipage}[h]{\linewidth}
		\centering 
		\includegraphics[width=\textwidth]{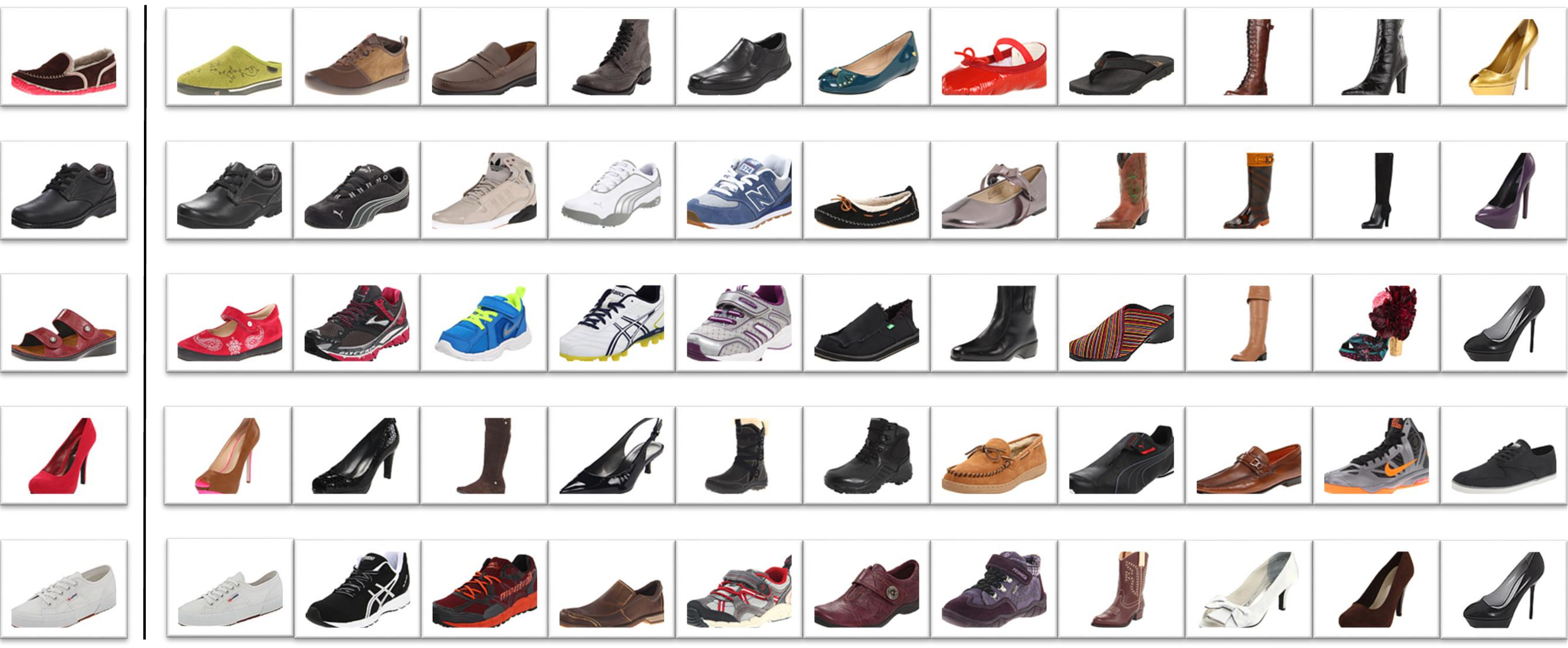}\\
		\mbox{({\it b}) {Height of Heels}}
	\end{minipage}
	\caption{Visualization of the image retrieval results for each of the two conditions (\ie, suggested gender and height of heels) on UT-Zappos-50k dataset with the learned embedding of {\name}$_{\rm Reg}$. The first image in each row is the query item, and shoes are ranked by distances to the query item in ascending order.}\label{fig:retrieval-2}
\end{figure*}

\paragraph{Visualization of conditional image retrieval.}
To gain insights into the conditions learned by our model ({\name}$_{\rm Reg}$), we provide image-retrieval visualizations for four conditions on UT-Zappos-50k in Fig.~\ref{fig:retrieval-1} and Fig.~\ref{fig:retrieval-2}. Generally speaking, {\name} can learn the distance between images based on a certain semantic. For example, on the ``suggested gender'' condition in Fig.~\ref{fig:retrieval-2} (a), our model can make all images related to Male (resp. Female) closer to the anchor related to Male (resp. Female) while pushing images related to Female (resp. Male) away.

\bibliographystyle{ieee_fullname}
\bibliography{Discover}